\documentclass{article}%
\usepackage{amsmath}
\usepackage{amsfonts}
\usepackage{amssymb}
\usepackage{graphicx}
\usepackage{algorithm}
\usepackage{appendix}
\usepackage{algpseudocode}%
\providecommand{\U}[1]{\protect \rule{.1in}{.1in}}
\newtheorem{theorem}{Theorem}
\newtheorem{proposition}[theorem]{Proposition}
\newtheorem{remark}{Remark}
\begin{document}

\title{Interpretable Ensembles of Hyper-Rectangles as Base Models}
\author{Andrei V. Konstantinov and Lev V. Utkin,\\Peter the Great St.Petersburg Polytechnic University\\St.Petersburg, Russia\\e-mail: andrue.konst@gmail.com, lev.utkin@gmail.com}
\date{}
\maketitle

\begin{abstract}
A new extremely simple ensemble-based model with the uniformly generated
axis-parallel hyper-rectangles as base models (HRBM) is proposed. Two types of
HRBMs are studied: closed rectangles and corners. The main idea behind HRBM is
to consider and count training examples inside and outside each rectangle. It
is proposed to incorporate HRBMs into the gradient boosting machine (GBM).
Despite simplicity of HRBMs, it turns out that these simple base models allow
us to construct effective ensemble-based models and avoid overfitting. A
simple method for calculating optimal regularization parameters of the
ensemble-based model, which can be modified in the explicit way at each
iteration of GBM, is considered. Moreover, a new regularization called the
"step height penalty" is studied in addition to the standard L1 and L2
regularizations. An extremely simple approach to the proposed ensemble-based
model prediction interpretation by using the well-known method SHAP is
proposed. 
It is shown that GBM with HRBM can be regarded as a model extending a set of interpretable models for explaining black-box models. 
Numerical experiments with real datasets illustrate the proposed GBM
with HRBMs for regression and classification problems. Experiments also
illustrate computational efficiency of the proposed SHAP modifications. The
code of proposed algorithms implementing GBM with HRBM is publicly available.

\textit{Keywords}: gradient boosting machine, SHAP, ensemble-based models,
explainability, rectangles

\end{abstract}

\section{Introduction}

Despite the rapid development of various approaches in machine learning, the
ensemble-based methodology remains one of the most effective approaches for
solving the regression and classification problems. Therefore, ensemble models
have been extensively studied in the machine learning community, and a huge
amount of methods for solving machine learning problems, including
classification and regression, have been developed in recent years
\cite{Xibin_Dong-etal-20,Ferreira-Figueiredo-2012,Jurek-etal-2014,Moreira-2012,Re-Valentini-2012,Ren-Zhang-Suganthan-2016,Rokach-2010,Sagi-Rokach-2018,Wozniak-etal-2014,ZH-Zhou-2012}%
. These methods are based on training a set of weak or base models from data
such that their predictions are combined in some way to obtain a strong
classifier or a regressor with a more accurate and generalizable result.

Two largest groups of ensemble-based methods can be pointed out. The first
group (bagging) consists of methods \cite{Breiman-1996} which are based on
constructing base models on subsets of training data. One of the best-known
bagging models is the Random Forest (RF) \cite{Breiman-2001}, which uses a
large number of randomly built individual decision trees, each trained on data
sets generated by means of bootstrap sampling. Another effective bagging model
is the Extremely Randomized Trees (ERTs), which is proposed by Geurts et al.
\cite{Geurts-etal-06}. In contrast to RFs, the ERT algorithm at each node
chooses a split point randomly for each feature and then selects the best
split among these features.

The second group (boosting) consists of methods based on a sequential and
dependent process. Well-known boosting models are AdaBoost
\cite{Freund-Shapire-97}, the gradient boosting machines (GBMs)
\cite{Friedman-2001,Friedman-2002} and its modifications XGBoost
\cite{Chen-Guestrin-2016}, LightGBM \cite{Guolin-etal-17}, CatBoost
\cite{Dorogush-etal-2018}. The main idea behind the gradient boosting methods
is to sequentially build each base model on the gradient descent direction of
a loss function, based on differences between the true values of target
variables and predicted values obtained from previous base models for all
training examples. Gradient boosting is typically used with decision trees as
base learners. Decision trees in gradient boosting provide accurate results in
reasonable computation time. Moreover, it is pointed out by Natekin and Knoll
\cite{Natekin-Knoll-13} that small trees in many practical results provide
better results, and there is much evidence that even complex models with rich
tree structure provide almost no benefit over compact trees. The assumption of
weak learners typically holds, and they can eventually generate a perfect fit.
In particular, a special case of a decision tree with only one split (a tree
stump) can be also successfully used in gradient boosting \cite{Jiang-02}.
Besides decision trees, other base-learner models can be incorporated into
gradient boosting \cite{Buhlmann-Hothorn-07}, including linear models
\cite{Buhlmann-06}, p-splines \cite{Schmid-Hothorn-08}, Markov random fields
\cite{Dietterich-etal-04}, wavelets \cite{Viola-Jones-01}.

In many applications, base models for boosting should be as simple as possible
to avoid overfitting caused by the greedy structure of algorithms. Following
the idea to simplify the base models, we propose an extremely simple model
that can be incorporated into a GBM as a base model for solving different
machine learning problems. The proposed model is called the
\emph{Hyper-Rectangle as the Base Model} (\emph{HRBM}). It is represented as
\emph{a closed axis-parallel rectangle or an axis-parallel rectangle corner}.
The main idea behind HRBM is to consider and count training examples inside
and outside each rectangle. From this point of view, the base models become to
be very simple. Rectangles are randomly generated at each iteration of GBM,
and the best rectangle among the generated ones is used as the base model at
the corresponding iteration. The best rectangle is determined in accordance
with a goal to cover at least one training point. As a result, a single
rectangle is used at each iteration of GBM. In spite of simplicity of HRBMs,
it turns out that these simple base models allow us to construct effective
ensemble-based models and avoid overfitting. GBM with HRBMs is denoted as
GBM-HRBM below.

An important property of HRBMs is that the corresponding ensemble-based models
can be simply interpreted by means of the interpretation method called SHapley
Additive exPlanations (SHAP) \cite{Lundberg-Lee-2017,Strumbel-Kononenko-2010},
which is inspired by game-theoretic Shapley values \cite{Shapley-1953}. SHAP
has an important shortcoming: its computational complexity depends on the
number of features and cannot be applied to models having data of the high
dimension without some approximation. Surprisingly, boosting models with HRBM
have a very simple interpretation in terms of SHAP. This is really an
interesting result. Moreover, GBM-HRBM due to its simple interpretation can be
regarded as an interpretable meta-model approximating a complex black-box model.

Two types of HRBMs are studied. The first one is based on uniformly generated
closed hyper-rectangles which have certain bounds. The second type is based on
uniformly generated corners.

Our contributions can be summarized as follows:

\begin{enumerate}
\item A new extremely simple ensemble-based model with axis-parallel
hyper-rectangles or corners as base models is proposed.

\item It is shown how HRBM incorporated into GBM. Simple expressions are
derived for searching optimal parameters of HRBMs at each iteration of the
boosting model and for computing the GBM-HRBM predictions.

\item A way for calculation of optimal regularization parameters of the
ensemble-based model is considered. The optimal regularization parameters can
simply be computed and modified at each iteration of GBM-HRBM. Moreover, a new
regularization called the \textquotedblleft step height
penalty\textquotedblright \ in addition to the standard $L_{1}$ and $L_{2}$
regularizations is proposed.

\item The problem of the GBM-HRBM prediction interpretation by using the
well-known method SHAP is solved in a computationally simple way. In fact, two
modifications of SHAP are proposed, which solves the interpretation problem in
a short time in comparison with using the original SHAP. The modifications are
called the model-based SHAP and the data-based SHAP. In accordance with the
first modification, the calculated Shapley values depend solely on the model
but non on the data whereas the second modification depends on the data.

\item Numerical experiments with well-known real datasets illustrate the
proposed GBM-HRBM for regression and classification problems. The experiments
also compare two types of the rectangle models: closed rectangles and corners.
Moreover, experiments with modifications of SHAP illustrate their
computational efficiency. The corresponding code implementing GBM-HRBM is
publicly available at: https://github.com/andruekonst/HRBM
\end{enumerate}

The paper is organized as follows. Related work can be found in Section 2. A
formal definition of HRBM is given in Section 3. Section 4 consider in detail
how HRBMs can be incorporated into GBM. Questions of regularization and its
optimal parameters are studied in Section 5. Algorithms of training and
testing GBM-HRBM are considered in Section 6. Modifications of SHAP for
interpretation predictions provided by ensemble-based models with HRBMs are
proposed in Section 7. Numerical experiments are provided in Section 8.
Concluding remarks can be found in Section 9. Appendix contains proofs of propositions.

\section{Related work}

\textbf{Ensemble-based models, GBM}. Ensemble-based methods can be regarded as
a powerful approach to improve predictive performance and robustness of
machine learning models. A detailed consideration of various types of
ensemble-based models can be found in Zhou's book \cite{ZH-Zhou-2012}.
Exhaustive descriptions of many ensemble approaches are presented in various
survey papers, for instance, in
\cite{Jurek-etal-2014,Kuncheva-2004,Mienye-Sun-22,Ren-Zhang-Suganthan-2016,Rokach-2019,Sagi-Rokach-2018,Wozniak-etal-2014}%
. Most authors assert that GBM \cite{Friedman-2001,Friedman-2002} and its
modifications, including XGBoost \cite{Chen-Guestrin-2016}, LightGBM
\cite{Guolin-etal-17}, CatBoost \cite{Dorogush-etal-2018}, can be viewed as
the best-known ensemble-based models.

Comprehensive surveys of GBMs and their comparison with other methods can be
found in
\cite{Bentejac-etal-21,Hancock-Khoshgoftaar-20,He-Lin-Lau-Wu-19,Mayr-etal-14}.
Many modifications of GBMs use decision trees as one of the most accurate base
models in GBMs \cite{Lundberg-etal-20,Natekin-Knoll-13}. The same can be said
about ERTs \cite{Konstantinov-Utkin-21a}. However, our study shows that HRBM
as the base model can lead to better models from the computation and accuracy
points of view.

\textbf{Interpretation methods. }Many methods have been developed to explain
black-box models. The first well-known method is the Local Interpretable
Model-agnostic Explanations (LIME) \cite{Ribeiro-etal-2016}. The main idea
behind the method is to approximate predictions of a black-box model by a
linear function of features. Due to success of LIME, various its modifications
have been also proposed
\cite{Garreau-Luxburg-2020a,Huang-Yamada-etal-2020,Kovalev-Utkin-Kasimov-20a,Rabold-etal-2019,Ribeiro-etal-2018}%
.

Another explanation method is the SHAP
\cite{Lundberg-Lee-2017,Strumbel-Kononenko-2010}. It is based on applying a
game-theoretic approach and Shapley values \cite{Shapley-1953}. Broeck et al.
\cite{Broeck-etal-21} studied general questions of the SHAP computational
efficiency. Various modifications of SHAP have been developed to explain
different machine learning models and tools
\cite{Aas-etal-2019,Antwarg-etal-20,Begley-etal-20,Bento-etal-20,Bouneder-etal-20,Takeishi-19,Yuan-Yu-20}%
. Applications of SHAP can be found in
\cite{Bi-Xiang-etal-20,Mangalathu-etal-20,Rodriguez-20}, Approaches to reduce
the computational complexity of SHAP were also proposed in
\cite{Benard-etal-21,Frye-etal-2020,Jethani-etal-21,Rozemberczki-Sarkar-21,Utkin-Konstantinov-22n}%
. Many interpretation methods and their comparison were considered and studied
in survey papers
\cite{Belle-Papantonis-2020,Guidotti-2019,Xie-Ras-etal-2020,Adadi-Berrada-2018,Arrieta-etal-2020,Carvalho-etal-2019,Das-Rad-20,Rudin-2019}
in detail.

Several interpretation approaches have been developed to efficiently explain
predictions of GBM by using SHAP
\cite{Amoukou-etal-22,Delgado-Panadero-etal-22,Futagami-etal-21,Loecher-etal-22,Lundberg-etal-18,Mayer-22,QingyaoSun-22,Teodoro-etal-23}%
. However, the proposed HRBMs incorporated into GBM make SHAP computationally
very simple by high-dimensional data.

\section{Formal definition of HRBM}

We introduce HRBM that can be incorporated into GBM as well as other ensemble
models as a base model for solving various machine learning problems.

Given $N$ training data (examples) $S=\{(\mathbf{x}_{1},y_{1}),...,(\mathbf{x}%
_{N},y_{N})\}$, in which each vector $\mathbf{x}_{i}=(x_{i}^{(1)}%
,...,x_{i}^{(d)})$ may belong to an arbitrary set $\mathcal{X}$ and represents
a feature vector involving $d$ features, $y_{i}\in \mathcal{Y}=\{1,...,C\}$
represents the class of the associated examples in the classification task or
$y_{i}\in \mathcal{Y}\subset \mathbb{R}$ represents the observed outputs in the
regression task. Machine learning aims to construct a classifier or a
regression model $f(\mathbf{x})$ that minimizes the expected risk
$\mathbb{E}_{(X.Y)\sim p(\mathbf{x},y)}\left[  l(Y,f(X))\right]  $, where
$p(\mathbf{x},y)$ is a joint density, and $l:\mathcal{Y}\times \mathcal{X}%
\rightarrow \mathbb{R}_{+}$ is a loss function.

Let $\mathbf{r}$ be a $d$-dimensional rectangle defined in the Cartesian space
$\mathbb{R}^{d}$ as
\begin{equation}
\mathbf{r}=\prod_{j=1}^{d}[a^{(j)},b^{(j)}],
\end{equation}
where $\mathbf{a}=(a^{(1)},...,a^{(d)})$ and $\mathbf{b}=(b^{(1)}%
,...,b^{(d)})$ are two vectors such that $a^{(j)}\leq b^{(j)}$ for all
$j=1,...,d$, which form the $d$-dimensional rectangle or hyper-rectangle.

Then HRBM is a piecewise-constant function with two distinct values: inside
and outside the rectangle. It can be represented as:
\begin{equation}
A(\mathbf{x};\mathbf{r},v_{in},v_{out})=\mathbb{I}[\mathbf{x}\in
\mathbf{r}](v_{in}-v_{out})+v_{out}, \label{APR-1}%
\end{equation}
where $v_{in}$ and $v_{out}$ are values inside and outside the rectangle which
are defined below; $\mathbb{I}[\mathbf{x}\in \mathbf{r}]$ is the indicator
function taking value $1$ if $\mathbf{x}\in \mathbf{r}$, and value $0$ if
$\mathbf{x}\notin \mathbf{r}$.

It follows from (\ref{APR-1}) that function $A$ takes two values:
$A(\mathbf{x};\mathbf{r},v_{in},v_{out})=v_{in}$ if point $\mathbf{x}$ belongs
to rectangle $\mathbf{r}$, and $A(\mathbf{x};\mathbf{r},v_{in},v_{out}%
)=v_{out}$, if $\mathbf{x}$ does not belong to $\mathbf{r}$.

Given a rectangle $\mathbf{r}$, values $v_{in}$ and $v_{out}$ can be found by
solving the following optimization problems:
\begin{equation}
v_{in}=\arg \min_{v}\sum_{i=1}^{N}l(y_{i},v)\cdot \mathbb{I}[\mathbf{x}_{i}%
\in \mathbf{r}], \label{APR-2}%
\end{equation}%
\begin{equation}
v_{out}=\arg \min_{v}\sum_{i=1}^{N}l(y_{i},v)\cdot \mathbb{I}[\mathbf{x}%
_{i}\notin \mathbf{r}]. \label{APR-3}%
\end{equation}

Here $l(y_{i},v)$ is a loss function which penalizes the distance between
$y_{i}$ and $v$. It can be seen from (\ref{APR-2}) and (\ref{APR-3}) that
value $v_{in}$ is nothing else but a simple prediction obtained on all
training points which fall into rectangle $\mathbf{r}$. For example, if we
take the regression loss $l(y_{i},v)=(y_{i}-v)^{2}$, then $v_{in}$ is the mean
value of observed outputs corresponding to $\mathbf{x}_{i}$ inside
$\mathbf{r}$. The same can be said about $v_{out}$ which is determined
similarly, but using all points $\mathbf{x}_{i}$ outside $\mathbf{r}$.

It is important to point out that values of $v_{out}$ can be replaced with the
bias term which is pre-calculated for the whole dataset as follows:%
\begin{equation}
q=\arg \min_{q}\sum_{i=1}^{N}l(y_{i},q).
\end{equation}

Hence, $v_{out}$ is computed as $v_{out}=q-v_{in}$. On the one hand, this
representation of rectangles is more effective from the computational point of
view because the bias $q$ is computed without analyzing whether points
$\mathbf{x}_{i}$ belong to a rectangle. On the other hand, we use the first
representation with $v_{in}$ and $v_{out}$ to simplify consideration of the
rectangle properties.

There are different types of rectangles which define their generation. We
consider the following two types:

\begin{enumerate}
\item \textbf{Closed rectangles:} when all elements of vectors $\mathbf{a}$
and $\mathbf{b}$ are finite. In this case, we can write
\begin{equation}
\mathbf{x}\in \mathbf{r}\iff \bigwedge_{j=1}^{d}\left(  x^{(j)}\in \lbrack
a^{(j)},b^{(j)}]\right)  .
\end{equation}

The rectangle can also be defined by taking its center $\mathbf{c}%
=(c^{(1)},...,c^{(d)})$ and its width vector $\mathbf{w}=\mathbf{b}%
-\mathbf{a}$.

\item \textbf{Corners of rectangles:} when some elements of vectors
$\mathbf{a}$ or $\mathbf{b}$ are unrestricted. In this case, we have a corner
of the rectangle. It can be defined by the corner center $\mathbf{c}$ and
conditions $a^{(j)}\rightarrow-\infty$ or $b^{(j)}\rightarrow \infty$.
\end{enumerate}

We also will denote the $j$-th rectangle feature (the edge) as $r^{(j)}%
=[a^{(j)},b^{(j)}]$.

It follows from the above introduced HRBM that it is a very simple model which
can be used as a base model in ensemble-based models. Therefore, we consider
how HRBM can be efficiently incorporated into ensembles, in particular, into GBM.

\section{Ensembles of HRBMs}

Inherently, HRBM is an extremely simple model, and thus cannot approximate
complex functions. However, ensembles of such models are much more expressive,
but only in a case when the base models are dependent. Indeed, consider a set
of independent HRBM models $A_{1}(\mathbf{x}),\dots,A_{k}(\mathbf{x})$. In
case of the mean squared loss function, values inside and outside each
rectangle are calculated as averages of target values for training points
which fall into the corresponding rectangle. By averaging the ensemble
predictions, we get for an example $\mathbf{x}$:
\begin{equation}
A(\mathbf{x})=\frac{1}{k}\sum_{i=1}^{k}A_{i}(\mathbf{x})=\frac{1}{k}\sum
_{i=1}^{k}(v_{out}^{(i)}+\mathbb{I}[\mathbf{x}\in \mathbf{r}_{i}](v_{in}%
^{(i)}-v_{out}^{(i)}))\approx \overline{y},
\end{equation}
which is close to constant $\overline{y}$.

Indeed, let us express $v_{out}^{(i)}$ through $q-v_{in}^{(i)}$. Then
\begin{align}
A(\mathbf{x})  &  =\frac{1}{k}\sum_{i=1}^{k}(q-v_{in}^{(i)}-\mathbb{I}%
[\mathbf{x}\in \mathbf{r}_{i}]q)\nonumber \\
&  =q\left(  1-\frac{k_{in}}{k}\right)  -\frac{1}{k}\sum_{i=1}^{k}v_{in}^{(i)}%
\end{align}
where $k_{in}=\sum_{i=1}^{k}\mathbb{I}[\mathbf{x}\in \mathbf{r}_{i}]$.

Note that $v_{in}^{(i)}$ under condition of the mean squared loss function can
be regarded as a mean $\overline{y}_{i}$ of target values corresponding to
points which fall into the $i$-th rectangle. If rectangles are independent and
random, then they correspond to bootstrapping that uses random sampling with
replacement. Hence, we can write
\begin{equation}
A(\mathbf{x})=q\left(  1-\frac{k_{in}}{k}\right)  -\widetilde{y}=\overline
{y}=\mathrm{const},
\end{equation}
where $\widetilde{y}$ is the mean of the bootstrap sample.

Taking into account the above, we propose to use GBM which makes rectangles dependent.

Below we use $\mathbb{I}_{in}^{(i)}$ and $\mathbb{I}_{out}^{(i)}$ as abridged
notations of $\mathbb{I}[\mathbf{x}_{i}\in \mathbf{r}]$ and $\mathbb{I}%
[\mathbf{x}_{i}\notin \mathbf{r}]$, respectively.

\subsection{HRBM in Gradient Boosting}

Let us consider GBM \cite{Friedman-2002} which can be regarded as an ensemble
method with \emph{dependent base models}. The main idea behind GBM is to
sequentially build each base model on the gradient descent direction of a loss
function, based on the residual (the difference between predicted value and
the true value of each example) from previous models \cite{Natekin-Knoll-13}.
In other words, GBM iteratively improves predictions $F_{i}(\mathbf{x})$ of
$y$ for $\mathbf{x}$ with respect to the so-called residual approximation loss
function by adding new base learners $f_{i}(\mathbf{x})\in \mathcal{F}$ that
improve upon the previous ones, forming an additive ensemble model of size
$T$:%
\begin{equation}
F_{0}(\mathbf{x})=\overline{y},\  \ F_{i}(\mathbf{x})=F_{i-1}(\mathbf{x}%
)+\gamma f_{i}(\mathbf{x}),\ i=1,...,T,
\end{equation}
where $\gamma$ is a learning rate; $\mathcal{F}$ is a set of base learners,
for instance, the set of decision trees; $\overline{y}$ is the mean target
value over the whole dataset, which is regarded as an initialization of boosting.

The algorithm aims to minimize a loss function $l$, for instance, the squared
error $L_{2}$-loss, by iteratively computing the gradient in accordance with
the standard gradient descent method. A single base model implementing
function $f_{i}$, for example, the decision tree, is constructed at each
iteration to fit the negative gradients. It is trained on a new dataset
$\{(\mathbf{x}_{j},\rho_{j}^{(i)})\}$, where $\rho_{j}^{(i)}$, $j=1,...,N$,
are residuals defined as partial derivatives of the expected loss function at
each point $\mathbf{x}_{j}$.

We propose to use HRBM as a base model in GBM, i.e., we take
\begin{equation}
f_{k}(\mathbf{x})=A(\mathbf{x};\theta_{k}),
\end{equation}
where the parameter vector $\theta_{k}$ is defined as
\begin{equation}
\theta_{k}=(\mathbf{r}_{k},v_{in},v_{out}).
\end{equation}

The parameter vector $\theta_{k}$ is chosen to optimize some functional. The
first approach for implementing that is to approximate the negative
residuals:
\begin{equation}
\rho_{i}^{(k)}=\left.  -\frac{\partial l(y_{i},z)}{\partial z}\right \vert
_{z=F_{k-1}(\mathbf{x}_{i})}.
\end{equation}

Hence, we can write%
\begin{equation}
\theta_{k}=\arg \min_{\theta_{k}}\sum_{i=1}^{N}\hat{l}\left(  A(\mathbf{x}%
_{i};\theta_{k}),\rho_{i}^{(k)}\right)  ,
\end{equation}
where $\hat{l}$ is the residual approximation loss function.

A more advanced approach is based on applying the second derivatives of the
loss function. Let us denote the first order derivatives as $g_{i}$:
\begin{equation}
g_{i}=\left.  \frac{\partial l(y_{i},z)}{\partial z}\right \vert _{z=F_{k-1}%
(\mathbf{x}_{i})},
\end{equation}
and the second order derivatives as $h_{i}$:
\begin{equation}
h_{i}=\left.  \frac{\partial^{2}l(y_{i},z)}{\partial z^{2}}\right \vert
_{z=F_{k-1}(\mathbf{x}_{i})}.
\end{equation}

It should be noted that $g_{i}$ and $h_{i}$ depend on $\mathbf{x}_{i}$ as well
as on the number $k$ of iteration. However, we omit the iteration index for
brevity because $v_{in}$ and $v_{out}$ are determined through $g_{i}$ and
$h_{i}$ for each iteration.

Note that we set $\gamma$ equal to $1$ when we build $f_{k}$, i.e., when we
are searching for optimal values $v_{in}$ and $v_{out}$ in order to further
reduce the absolute value of HRBM by multiplying by $\gamma$. The loss
function can be expanded as follows:
\begin{equation}
l(y_{i},F_{k}(\mathbf{x}_{i}))=l(y_{i},F_{k-1}(\mathbf{x}_{i}))+g_{i}\cdot
f_{k}(\mathbf{x}_{i})+\frac{1}{2}h_{i}(f_{k}(\mathbf{x}_{i}))^{2}%
+o((f_{k}(\mathbf{x}_{i}))^{2}), \label{APR-8}%
\end{equation}
so the empirical loss minimization is equivalent to the following optimization
problem:
\begin{equation}
\frac{1}{N}\sum_{i=1}^{N}\left(  g_{i}\cdot f_{k}(\mathbf{x}_{i})+\frac{1}%
{2}h_{i}(f_{k}(\mathbf{x}_{i}))^{2}+o((f_{k}(\mathbf{x}_{i}))^{2})\right)
\rightarrow \min.
\end{equation}

In case of small values $f_{k}(\mathbf{x}_{i})$, the optimization problem can
be approximately rewritten as follows:
\begin{equation}
\hat{L}=\frac{1}{N}\sum_{i=1}^{N}\left(  g_{i}\cdot f_{k}(\mathbf{x}%
_{i})+\frac{1}{2}h_{i}(f_{k}(\mathbf{x}_{i}))^{2}\right)  \rightarrow \min.
\label{eq:quardatic_loss_expansion}%
\end{equation}

We have two different base algorithms. If we would not know function $f$,
i.e., we have a model-agnostic gradient boosting, then problem
(\ref{eq:quardatic_loss_expansion}) leads to residuals of form:%
\begin{equation}
\rho_{i}=-\frac{g_{i}}{h_{i}}.
\end{equation}

The optimal HRBM values can be found in this case as follows:%
\begin{equation}%
\begin{cases}
v_{in}=-\frac{1}{\sum_{i=1}^{N}\mathbb{I}_{in}^{(i)}}\sum_{i=1}^{N}%
\mathbb{I}_{in}^{(i)}\cdot \dfrac{g_{i}}{h_{i}},\\
v_{out}=-\frac{1}{\sum_{i=1}^{N}\mathbb{I}_{out}^{(i)}}\sum_{i=1}%
^{N}\mathbb{I}_{out}^{(i)}\cdot \dfrac{g_{i}}{h_{i}}.
\end{cases}
\label{APR-19}%
\end{equation}

Since the base model is HRBM, then a more precise solution can be obtained.
First, for the sake of brevity, we define the function $f_{k}(\mathbf{x}_{i})$
for the $k$-th current iteration as:
\begin{equation}
f_{k}(\mathbf{x}_{i})=\mathbb{I}_{in}^{(i)}v_{in}+\mathbb{I}_{out}%
^{(i)}v_{out}. \label{APR-20}%
\end{equation}

Let us denote also for brevity:%
\begin{equation}
G_{in}=\sum_{i=1}^{N}\mathbb{I}_{in}^{(i)}\cdot g_{i},\ G_{out}=\sum_{i=1}%
^{N}\mathbb{I}_{out}^{(i)}\cdot g_{i}, \label{APR-20-1}%
\end{equation}%
\begin{equation}
H_{in}=\sum_{i=1}^{N}\mathbb{I}_{in}^{(i)}\cdot h_{i},\ H_{out}=\sum_{i=1}%
^{N}\mathbb{I}_{out}^{(i)}\cdot h_{i}. \label{APR-20-2}%
\end{equation}

The loss function (\ref{eq:quardatic_loss_expansion}) has a minimum at a point
with zero derivatives, namely:
\begin{equation}
\dfrac{\partial \hat{L}}{\partial v_{in}}=\dfrac{1}{N}\left(  G_{in}%
+H_{in}\cdot v_{in}\right)  =0,
\end{equation}%
\begin{equation}
\dfrac{\partial \hat{L}}{\partial v_{out}}=\dfrac{1}{N}\left(  G_{out}%
+H_{out}\cdot v_{out}\right)  =0.
\end{equation}

It follows from the above that optimal values of $v_{in}$ and $v_{out}$ can be
calculated from the following simple expressions:%
\begin{equation}
v_{in}=-\frac{G_{in}}{H_{in}},\ v_{out}=-\frac{G_{out}}{H_{out}}.
\label{APR-22}%
\end{equation}

Optimal values of $v_{in}$ and $v_{out}$ obtained by using (\ref{APR-19}) and
(\ref{APR-22}) are different in general. For the loss functions with a
constant second derivative, such as mean squared error, the values
(\ref{APR-19}) and (\ref{APR-22}) are the same, but, for more complex
functions, such as the cross-entropy, solution (\ref{APR-22}) is more
accurate. In addition, the second solution, in contrast to the classical
gradient boosting, allows us to introduce regularization on the model values
$v_{in}$ and $v_{out}$.

If we multiply the base functions by the learning rate $\gamma$, then the
result only indirectly control the model smoothness and has different effect
which depend on absolute values of the loss function gradient. At that, values
$v_{in}$ and $v_{out}$ are of the form:%
\begin{equation}
v_{in}=-\gamma \frac{G_{in}}{H_{in}},\ v_{out}=-\gamma \frac{G_{out}}{H_{out}}.
\end{equation}

A rectangle is a weak model, but if it is trained on a small number of
training sample points, then the model may also lead to overfitting. In
addition, noise in the target variable also has a significant effect on
increase of the error. To overcome these difficulties, the gradient boosting
algorithm can be improved by estimating the accuracy of base models on a
validation set. To implement that, a base model is built based on the training
set at each iteration. Then it is verified whether it does not reduce the
accuracy on the validation set. Finally, if the validation is successful, then
values $v_{in}$ and $v_{out}$ are recalculated on the entire training set. In
this case, the division into training and validation sets can be performed
once as well as at each iteration.%

\begin{figure}
[ptb]
\begin{center}
\includegraphics[
height=2.0324in,
width=4.9347in
]%
{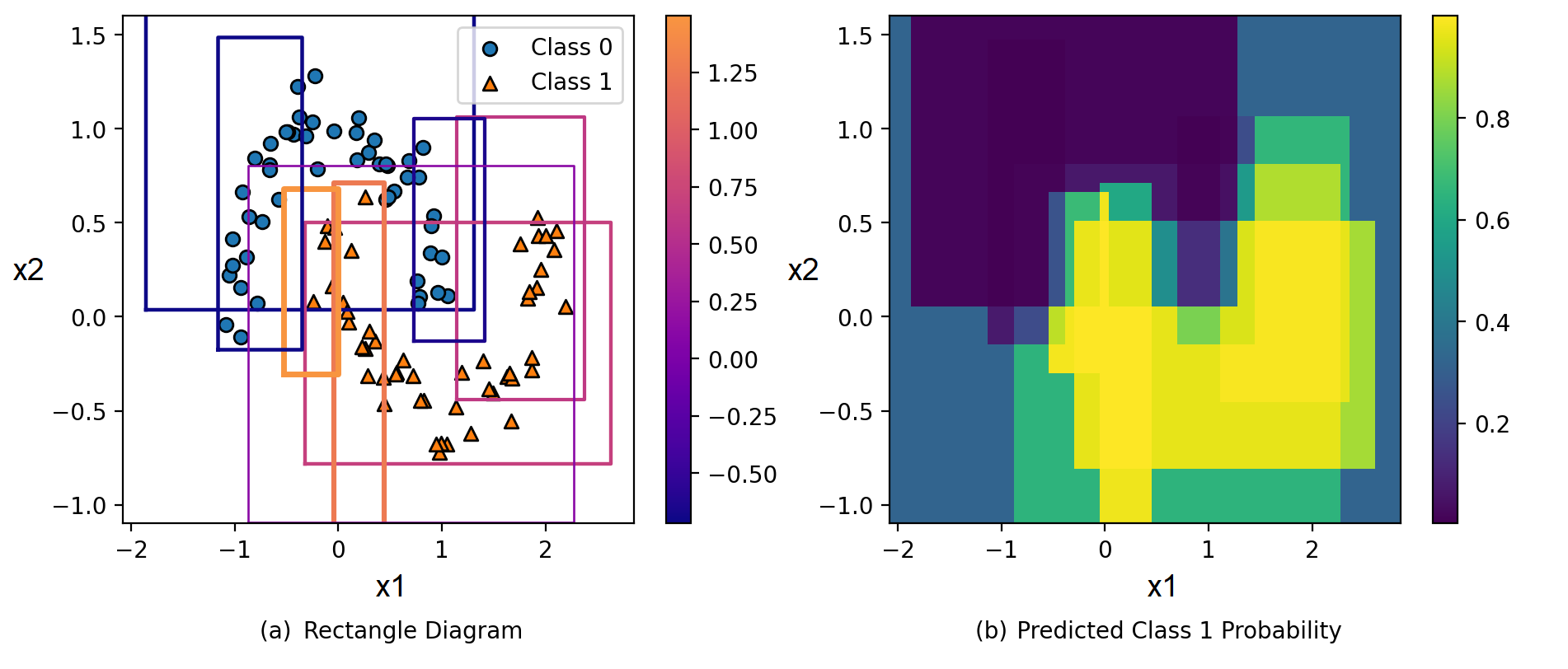}%
\caption{An illustrative example of HRBMs in the form of rectangles in
classification on the two moons dataset: (a) generated rectangles after $8$
iterations; (b) predicted probabilities of Class 1 obtained by using GBM-HRBM
with $8$ iterations}%
\label{f:clf_rectangles_bce2}%
\end{center}
\end{figure}

An illustrative classification example using the well-known \textquotedblleft
two moons\textquotedblright \ dataset is depicted in Fig.
\ref{f:clf_rectangles_bce2}. Fig. \ref{f:clf_rectangles_bce2} (a) shows closed
rectangles generated at each iteration of GBM-HRBM. Colors and thickness of
the rectangle lines correspond to $v_{in}$ for each rectangle, whose values
are logits of the class 1. For clarity, the model contains only 8 rectangles.
The loss function in this case is a binary cross entropy applied to the
sigmoid $\sigma$ from the gradient boosting output:
\begin{equation}
l(y,z)=y\cdot \ln(\sigma(z))+(1-y)\cdot \ln(1-\sigma(z)).
\end{equation}

Fig. \ref{f:clf_rectangles_bce2} (b) shows predicted probabilities of Class 1
obtained by using GBM-HRBM after $8$ iterations with $8$ rectangles shown in
Fig. \ref{f:clf_rectangles_bce2} (a).%

\begin{figure}
[ptb]
\begin{center}
\includegraphics[
height=3.1837in,
width=3.0262in
]%
{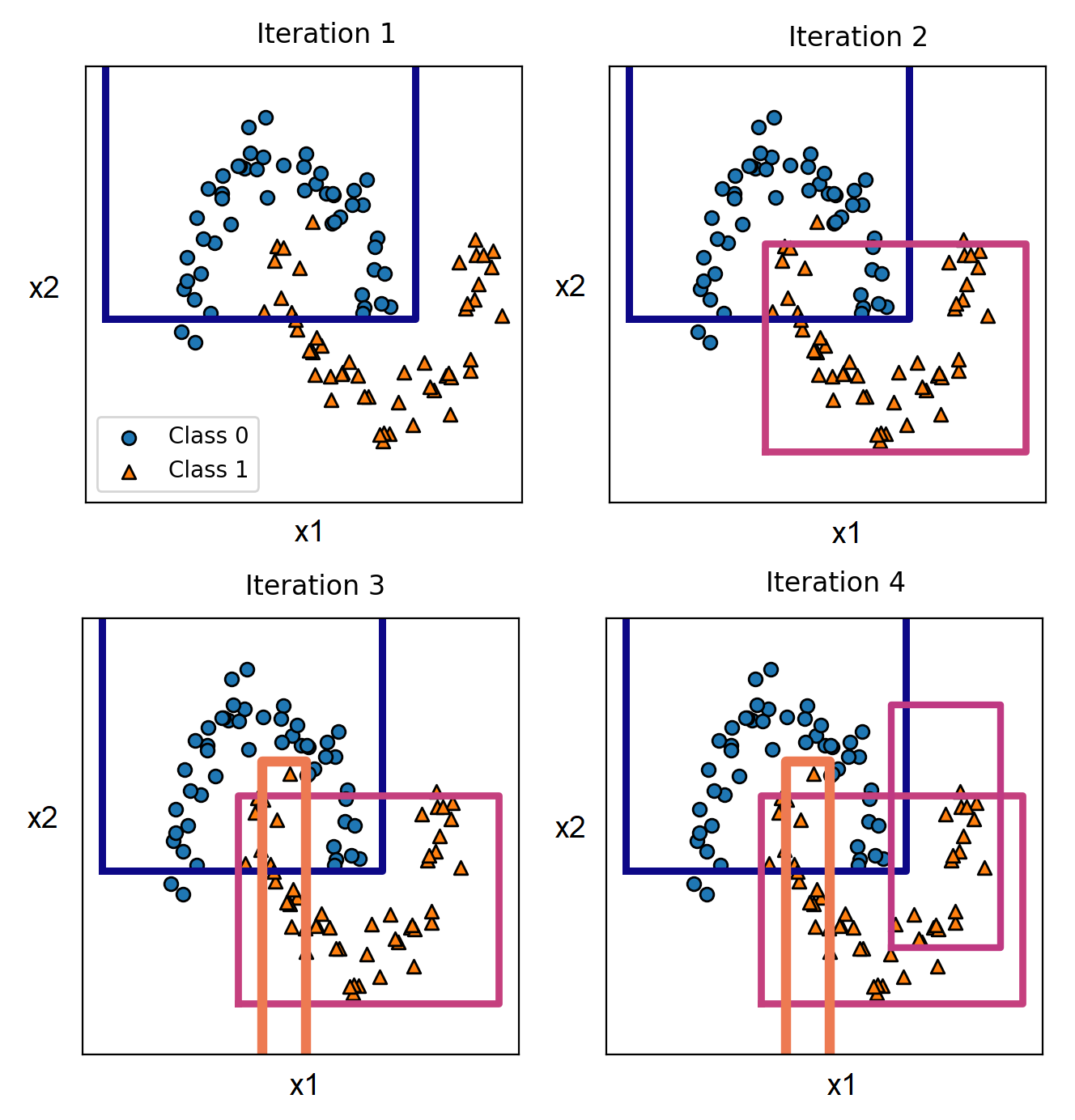}%
\caption{Rectangles generated at four iterations for classification}%
\label{f:sequence_class}%
\end{center}
\end{figure}

Fig. \ref{f:sequence_class} illustrates how the rectangles are generated at
each iteration. We show only four iterations. It can be seen from Fig.
\ref{f:sequence_class} that rectangles are generated to cover the data domain.%

\begin{figure}
[ptb]
\begin{center}
\includegraphics[
height=1.9014in,
width=4.4299in
]%
{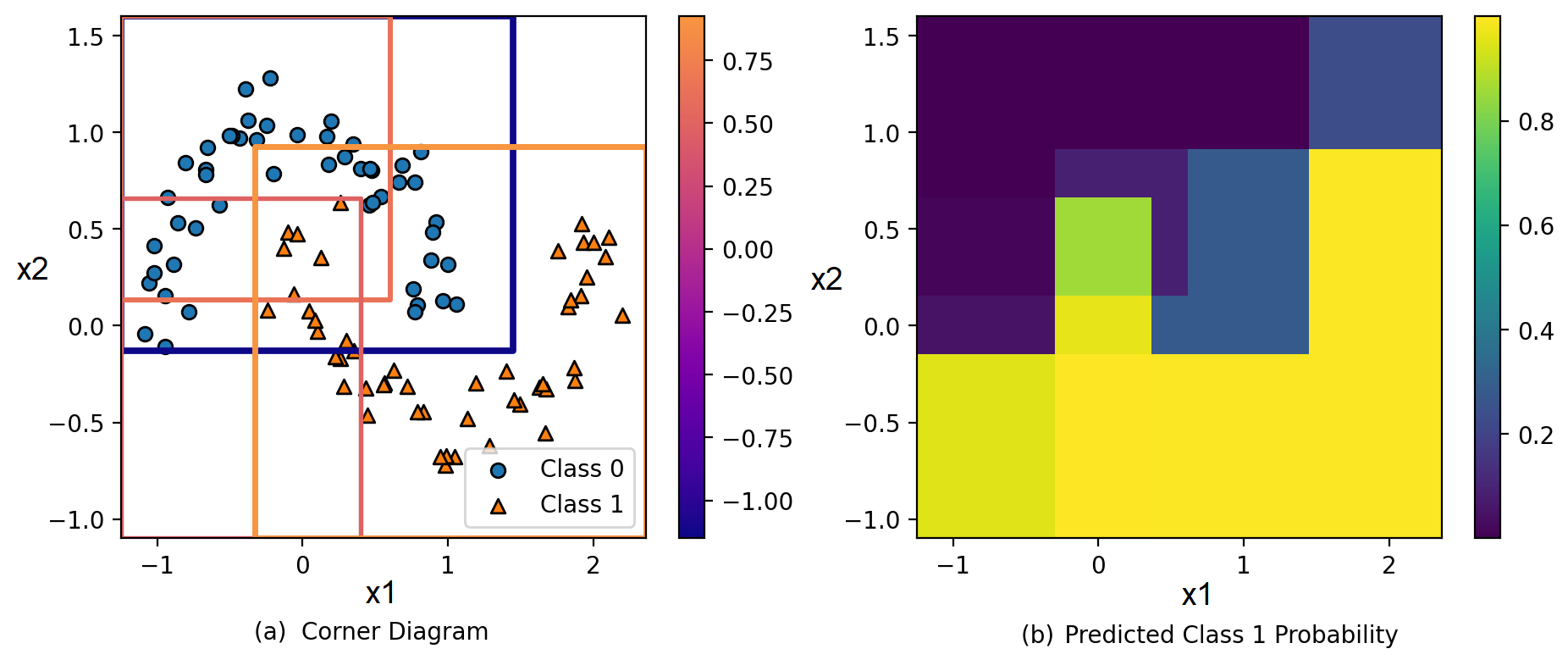}%
\caption{An illustrative example of HRBMs in the form of corners in
classification on the two moons dataset: (a) generated rectangles after $4$
iterations; (b) predicted probabilities of Class 1 obtained by using GBM-HRBM
with $4$ iterations}%
\label{f:clf_corners}%
\end{center}
\end{figure}

\begin{figure}
[ptb]
\begin{center}
\includegraphics[
height=3.3578in,
width=3.1494in
]%
{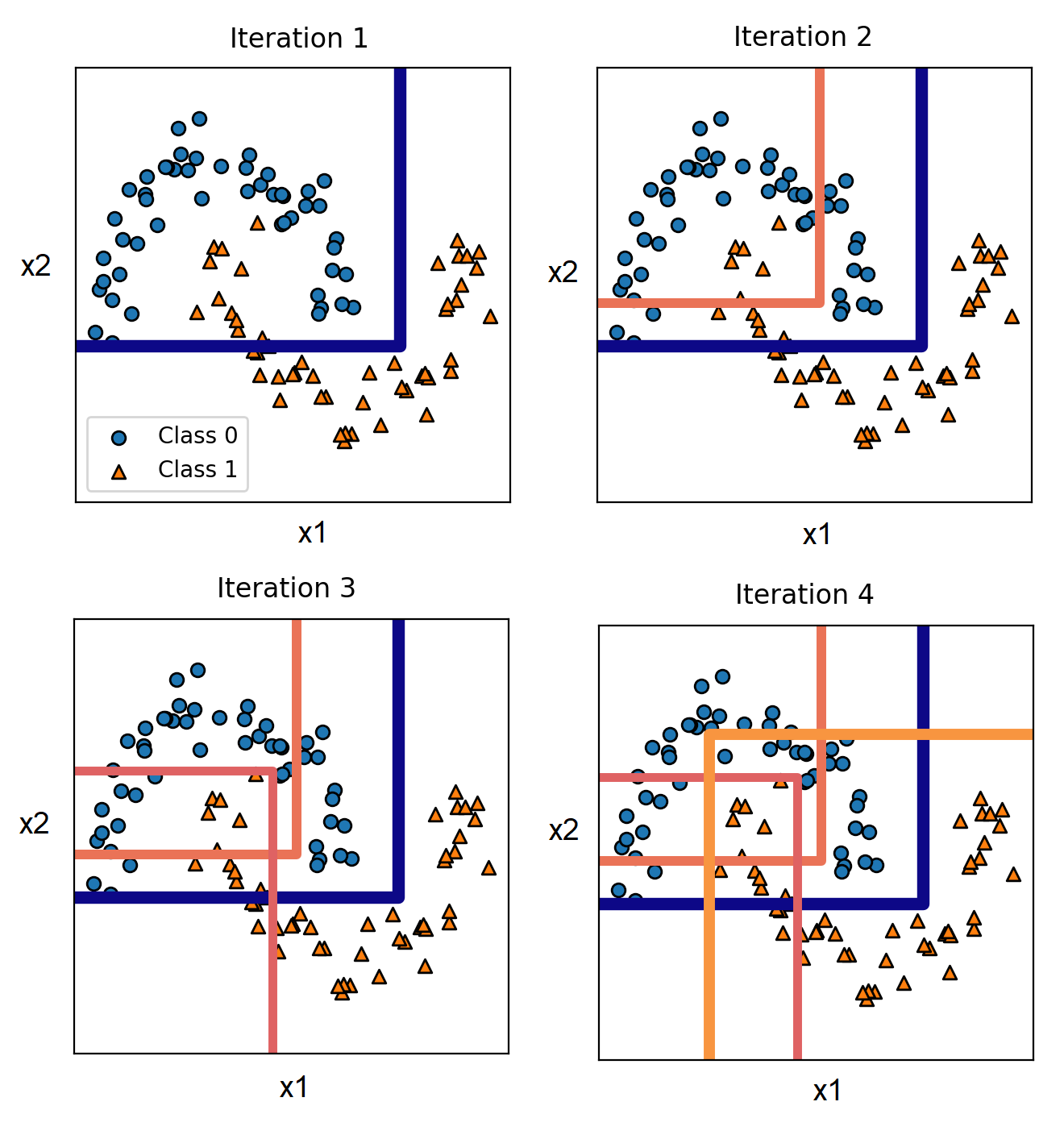}%
\caption{Corners generated at four iterations for classification}%
\label{f:clf_corners_sequence}%
\end{center}
\end{figure}

Fig. \ref{f:clf_corners} is similar to Fig. \ref{f:clf_rectangles_bce2}, but
it illustrates classification example using $4$ generated corners instead of
$8$ rectangles. In particular, Fig. \ref{f:clf_corners} (a) shows $4$ corners
generated during $4$ iterations of GBM-HRBM. Fig. \ref{f:clf_corners} (b)
shows predicted probabilities of Class 1 obtained by using GBM-HRBM after $4$
iterations with $4$ corners shown in Fig. \ref{f:clf_corners} (a). We see that
the predicted probabilities are more uncertain in comparison with the case of
rectangles shown in Fig. \ref{f:clf_rectangles_bce2} (b). However, it does not
mean that corners provide worse results. We take only four corners (compare
with eight rectangles in the example in Fig. \ref{f:clf_rectangles_bce2}) in
order to make pictures with corners visible. Fig. \ref{f:clf_corners_sequence}
illustrates how the corners are generated at each iteration. We again show
only four iterations of GBM-HRBM.%

\begin{figure}
[ptb]
\begin{center}
\includegraphics[
height=2.0914in,
width=4.1484in
]%
{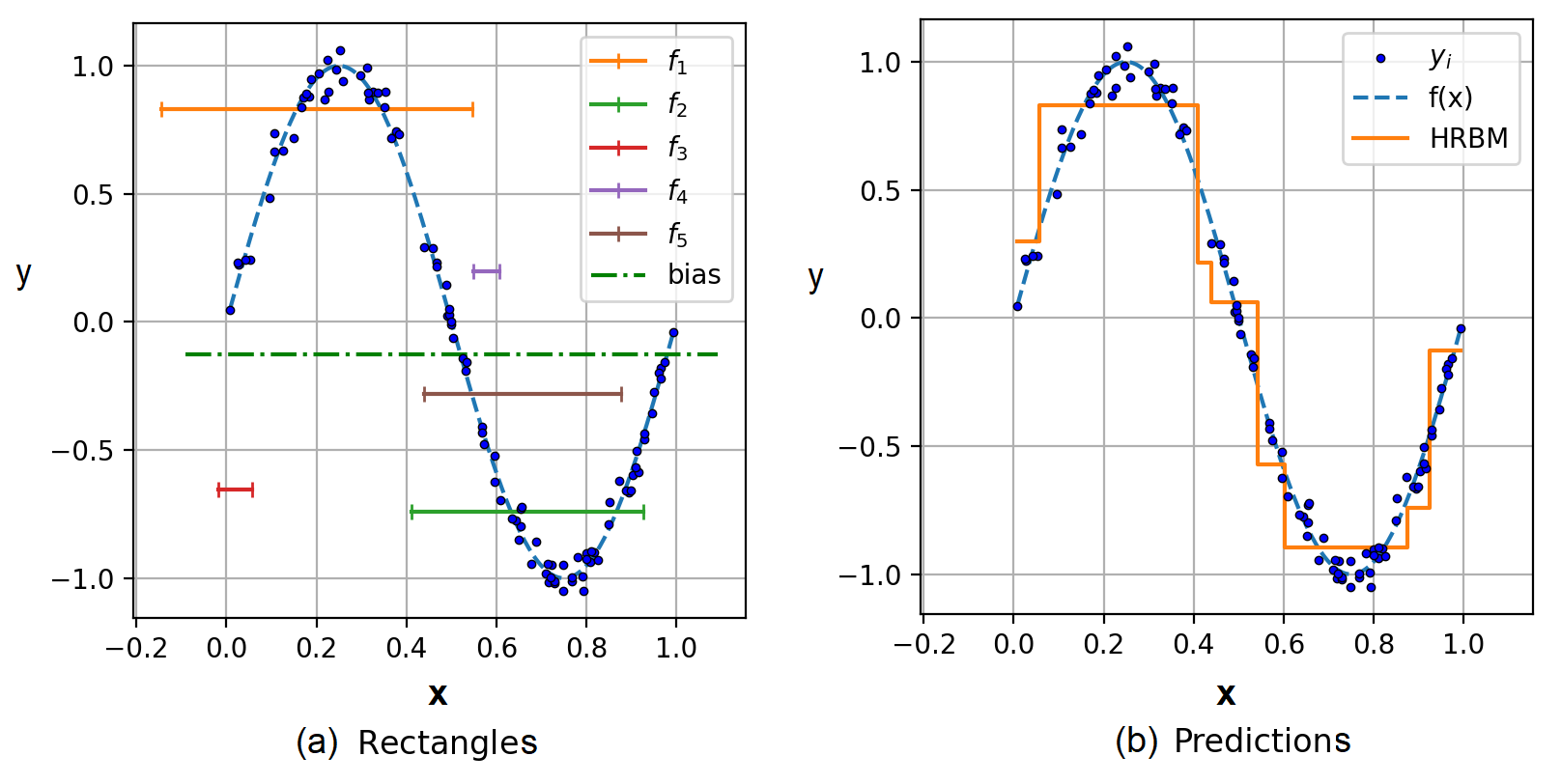}%
\caption{An illustrative example of HRBMs in regression: (a) generated
one-dimensional rectangles after $5$ iterations; (b) predictedregression
values obtained by using GBM-HRBM with $5$ iterations}%
\label{f:regres_rect}%
\end{center}
\end{figure}

\begin{figure}
[ptb]
\begin{center}
\includegraphics[
height=3.9699in,
width=3.9646in
]%
{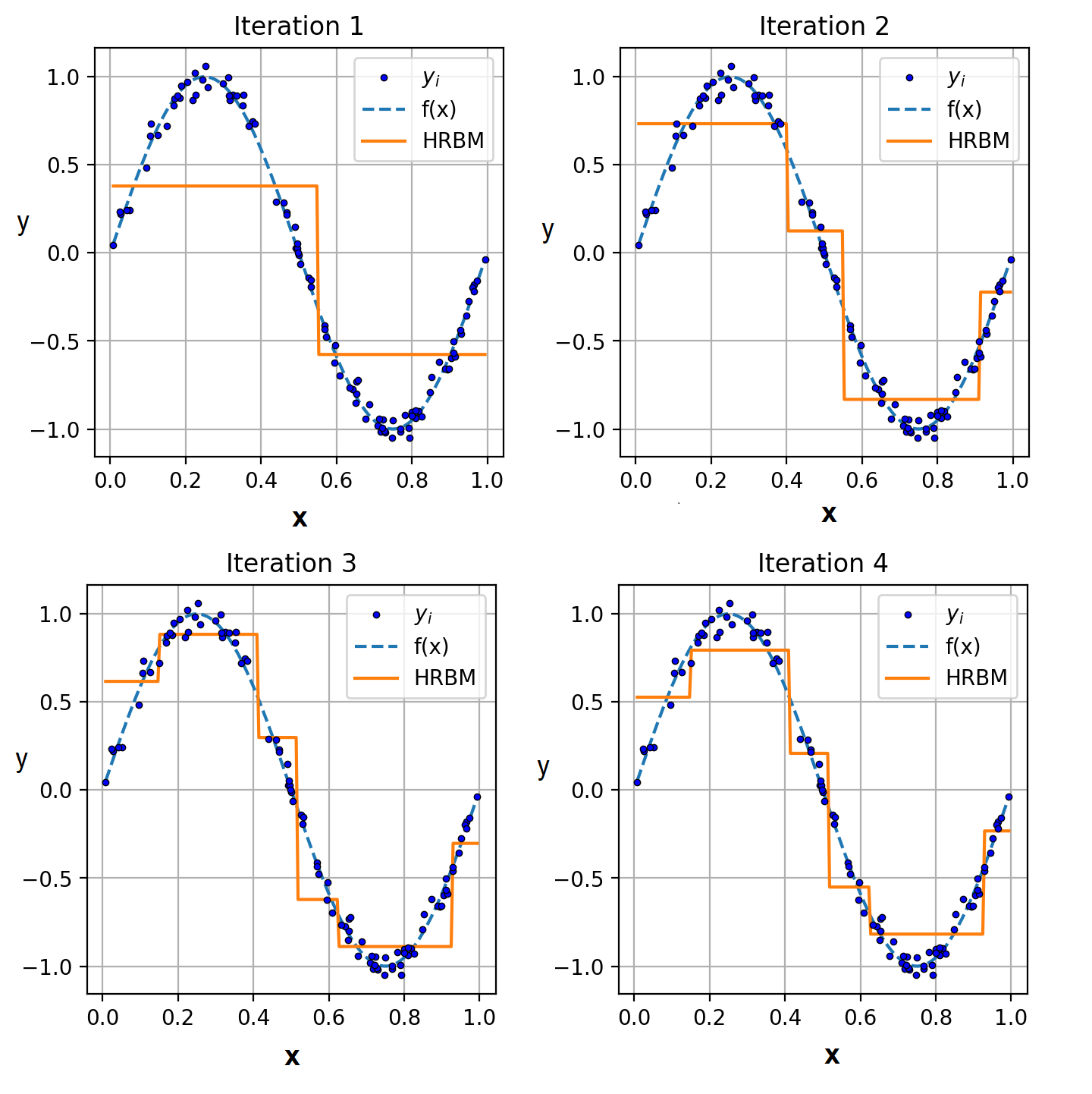}%
\caption{Rectangles generated at four iterations for regression}%
\label{f:regres_rect_iter}%
\end{center}
\end{figure}

Fig. \ref{f:regres_rect} illustrates an one-dimensional regression task
implemented by using GBM-HRBM with $5$ iterations ($5$ rectangles). Rectangles
in this case are represented by segments. Points (small circles) of the
training set, the unknown truth function (the dashed line), and rectangles
(segments) are depicted in Fig. \ref{f:regres_rect} (a). Each segment is
located at a height which is equal to the inside value $v_{in}$ of the
corresponding rectangle. Bias $q$ is depicted by the dash-and-dot line. Fig.
\ref{f:regres_rect} (b) illustrates predictions obtained by using GBM-HRBM
with 5 iterations. It can be seen from Fig. \ref{f:regres_rect} (b) that even
five iterations allow us to get accurate approximation of the unknown truth
function. Fig. \ref{f:regres_rect_iter} illustrates how rectangles in the
regression task are added at each iteration of GBM-HRBM. It can be seen from
Fig. \ref{f:regres_rect_iter} (Iteration 1) that the first rectangle ($f_{1}$
in Fig. \ref{f:regres_rect} (a)) divides all points into two subsets: (1)
points $\mathbf{x}$ which fall inside the corresponding segment $[0,0.5]$; (2)
points which fall outside the segment $[0.5,1]$. The second iteration
generates the rectangle $[0.4,0.92]$ ($f_{2}$ in Fig. \ref{f:regres_rect}
(a)). Its intersection with the first segment forms additional subsets of
points. It is interesting to see that rectangles after the second iteration
approximate the unknown function. The same process of adding rectangles is
depicted in other pictures of Fig. \ref{f:regres_rect_iter}.%

%

\section{Regularization}

In order to improve GBM-HRBM and avoid overfitting, we propose to use
regularization. First, we consider the standard $L_{1}$ and $L_{2}$
regularization. Let us add the regularization terms $\Omega(v_{in},v_{out})$
to the loss function $\hat{L}$ defined in (\ref{eq:quardatic_loss_expansion}).
A new loss function $\tilde{\mathcal{L}}$ is defined as%
\begin{equation}
\tilde{\mathcal{L}}=\hat{L}+\Omega(v_{in},v_{out})=\hat{L}+\lambda
_{1}\left \vert \vec{v}\right \vert +\frac{\lambda_{2}}{2}\lVert \vec{v}%
\rVert^{2} \label{APR-40}%
\end{equation}
where $\vec{v}=(v_{in},v_{out})^{\mathrm{T}}$; $\lambda_{1}$ and $\lambda_{2}$
are hyperparameters which control the strength of the standard $L_{1}$ and
$L_{2}$ regularizations, respectively; $\hat{L}$ is the loss function without
regularization defined in (\ref{eq:quardatic_loss_expansion}).

It is simply to get optimal values of $v_{in},v_{out}$ under condition of
adding the $L_{1}$ and $L_{2}$ regularization terms. We again use notations
introduced in (\ref{APR-20-1}) and (\ref{APR-20-2}).

\begin{proposition}
\label{prop:regular1}Suppose that the loss function with regularization is of
the form (\ref{APR-40}). If $\lambda_{1}=0$ (the $L_{2}$ regularization), then
optimal values of $v_{in}(\lambda_{2})$ and $v_{out}(\lambda_{2})$ as
functions of $\lambda_{2}$ are of the form:
\begin{equation}
v_{in}(\lambda_{2})=-\frac{G_{in}}{N\cdot \lambda_{2}+H_{in}},\ v_{out}%
(\lambda_{2})=-\frac{G_{out}}{N\cdot \lambda_{2}+H_{out}}. \label{APR-30}%
\end{equation}

If $\lambda_{1}>0$ and $\lambda_{2}>0$, then the optimal value of
$v_{in}(\lambda_{1},\lambda_{2})$ is determined as
\begin{equation}
v_{in}(\lambda_{1},\lambda_{2})=\left \{
\begin{array}
[c]{cc}%
-\frac{G_{in}+N\cdot \lambda_{1}}{N\cdot \lambda_{2}+H_{in}}, & G_{in}%
<-N\cdot \lambda_{1},\\
0, & \left \vert G_{in}\right \vert \leq N\cdot \lambda_{1},\\
-\frac{G_{in}-N\cdot \lambda_{1}}{N\cdot \lambda_{2}+H_{in}}, & G_{in}%
>N\cdot \lambda_{1}.
\end{array}
\right.  \label{APR-32}%
\end{equation}

$v_{out}(\lambda_{1},\lambda_{2})$ is written similarly if we replace
\textquotedblleft in\textquotedblright \ with \textquotedblleft
out\textquotedblright \ in (\ref{APR-32}).
\end{proposition}

Let us introduce a new type of the regularization function which is called the
\textquotedblleft step height penalty\textquotedblright. It can be expressed
as the distance between $v_{in}$ and $v_{out}$. The idea behind this type of
regularization is to minimize the distance between $v_{in}$ and $v_{out}$. In
this case, we write the regularization function as
\begin{equation}
\tilde{\mathcal{L}}=\hat{L}+\Omega_{h}(v_{in},v_{out})=\hat{L}+\eta
_{1}\left \vert v_{in}-v_{out}\right \vert +\frac{\eta_{2}}{2}\lVert
v_{in}-v_{out}\rVert^{2}. \label{APR-48}%
\end{equation}

Here $\eta_{1}$ and $\eta_{2}$ are hyperparameters which control the strength
of the $L_{1}$ and $L_{2}$ regularization terms, respectively.

\begin{proposition}
\label{prop:regular2}Suppose that the loss function with regularization is of
the form (\ref{APR-48}). If $\eta_{1}=0$ (the $L_{2}$-regularization), then
optimal values of $v_{in}(\eta_{2})$ and $v_{out}(\eta_{2})$ as functions of
$\eta_{2}$ are of the form:%
\begin{equation}
v_{in}(\eta_{2})=-\frac{G_{in}+N\eta_{2}G/H_{out}}{H_{in}+N\eta_{2}H/H_{out}},
\label{APR-52}%
\end{equation}%
\begin{equation}
v_{out}(\eta_{2})=-\frac{G_{out}+N\eta_{2}G/H_{in}}{H_{otu}+N\eta_{2}H/H_{in}%
}, \label{APR-52-1}%
\end{equation}
where
\begin{equation}
G=\sum_{i=1}^{N}g_{i},\ H=\sum_{i=1}^{N}h_{i}.
\end{equation}

Case $\eta_{1}>0$ does not have a sense because the same regularization can be
implemented by using only the case $\eta_{2}>0$ and $\eta_{1}=0$.
\end{proposition}

It follows from (\ref{APR-52}) that values of $v_{in}$ and $v_{out}$ by
$\eta_{2}>0$ and $\eta_{1}=0$ coincide by large values of $\eta_{2}$, and they
are
\begin{equation}
\lim_{\eta_{2}\rightarrow+\infty}v_{in}(\eta_{2})=\lim_{\eta_{2}%
\rightarrow+\infty}v_{out}(\eta_{2})=\frac{G}{H}.
\end{equation}

The above result can be interpreted as follows. If optimal values of $v_{in}$
and $v_{out}$ (without regularization) are far from each other, then the
addition of regularization shifts values of $v_{in}$ and $v_{out}$ towards
each other.

\subsection{Optimal regularization parameters}

To ensure the correctness of the approximation of the loss function obtained
by means of series expansion (\ref{APR-8}), as well as to reduce overfitting,
it is proposed to restrict the absolute values of function $f_{k}(\mathbf{x}%
)$. The first way implementing the restriction is to multiply function
$f_{k}(\mathbf{x})$ by the learning rate $\gamma$. The second way is based on
regularization which has been studied in Propositions \ref{prop:regular1} and
\ref{prop:regular2}.

First, we consider the standard $L_{1}$ and $L_{2}$ regularization
(\ref{APR-40}). Values of the regularization parameters $\lambda_{1}$ and
$\lambda_{2}$ can be estimated by bounding the HRBM absolute values by some
constant $\beta$ which can be regarded as an analogue of the learning rate
that limits each rectangle:%
\begin{equation}
\left \vert f_{k}(\mathbf{x})\right \vert \leq \beta. \label{eq:abs_le_beta}%
\end{equation}

The following proposition establishes the relationship between parameters
$\lambda_{1}$, $\lambda_{2}$ and $\beta$ when the $L_{1}$ or $L_{2}$
regularization is used.

\begin{proposition}
\label{prop:regular3}Suppose that the loss function with regularization is of
the form (\ref{APR-40}), and the condition (\ref{eq:abs_le_beta}) is
fulfilled. Then the optimal value $\underline{\lambda}_{2}$ of $\lambda_{2}$
as a function of $\beta$ is defined as follows:
\begin{equation}
\underline{\lambda}_{2}=\max \left(  {\underline{\lambda}_{2}^{in}%
,\underline{\lambda}_{2}^{out},0}\right)  , \label{APR-56}%
\end{equation}
where
\begin{equation}
\underline{\lambda}_{2}^{in}=\frac{1}{N}\left[  \frac{1}{\beta}\left \vert
G_{in}\right \vert -H_{in}\right]  , \label{APR-57}%
\end{equation}%
\begin{equation}
\underline{\lambda}_{2}^{out}=\frac{1}{N}\left[  \frac{1}{\beta}\left \vert
G_{out}\right \vert -H_{out}\right]  . \label{APR-58}%
\end{equation}

The optimal value $\overline{\lambda}_{1}$ of $\lambda_{1}$ as a function of
$\beta$ is defined as follows:
\begin{equation}
\overline{\lambda}_{1}=\min(\overline{\lambda}_{1}^{in},\overline{\lambda}%
_{1}^{out}). \label{APR-59}%
\end{equation}
where
\begin{equation}
\overline{\lambda}_{1}^{in}=\frac{1}{N}[\beta H_{in}-|G_{in}|],
\label{APR-59-1}%
\end{equation}%
\begin{equation}
\overline{\lambda}_{1}^{out}=\frac{1}{N}[\beta H_{out}-|G_{out}|].
\label{APR-59-2}%
\end{equation}

\end{proposition}

By using the above approach to choosing $\lambda_{1}$ or $\lambda_{2}$, we can
state that one of the values of $v_{in}$ or $v_{out}$ is equal to $\beta$
(except for the case when the first derivative is zero for the entire dataset,
i.e., $\forall i:~g_{i}=0$). Then the second value does not exceed $\beta$.

Let us consider again the condition (\ref{eq:abs_le_beta}) under condition
that the regularization parameters $\eta_{1}$ and $\eta_{2}$ are defined by
the step height penalty (\ref{APR-48}). First, we study the case when
$\eta_{1}=0$.

\begin{proposition}
\label{prop:regular4}Suppose that the loss function with regularization is of
the form (\ref{APR-48}), and condition (\ref{eq:abs_le_beta}) is fulfilled.
Denote
\begin{equation}
C_{1}=\frac{H_{out}}{N}\frac{G_{in}+\beta H_{in}}{G+\beta H},\ C_{2}%
=-\frac{H_{out}}{N}\frac{G_{in}+\beta H_{in}}{G-\beta H},
\end{equation}%
\begin{equation}
{B_{1}=-\frac{H_{in}}{N}\frac{G_{out}+\beta H_{out}}{G+\beta H},\ }%
B_{2}={\frac{H_{out}}{N}\frac{-G_{in}+\beta H_{in}}{G-\beta H},}%
\end{equation}%
\begin{equation}
{B_{3}=\frac{H_{in}}{N}\frac{-G_{out}+\beta H_{out}}{G-\beta H}.}%
\end{equation}
Then the optimal value $\underline{\eta}_{2}$ of $\eta_{2}$ as a function of
$\beta$ for $v_{in}$ is defined depending on three cases as follows:

If $G>\beta H$, then $\underline{\eta}_{2}=C_{1}$ for $v_{in}$ and
$\underline{\eta}_{2}=\max \left(  {0,-C}_{1}{,-B_{1}}\right)  $ for $v_{out}$.

If $|G|<\beta H$, then $\underline{\eta}_{2}=\max \left(  C_{2}{,}C_{1}\right)
$ for $v_{in}$ and $\underline{\eta}_{2}=\max({0,B_{1},B_{2},{B}_{3},C_{2})}$
for $v_{out}$.

If $G<-\beta H$, then $\underline{\eta}_{2}=C_{2}$ for $v_{in}$ and
$\underline{\eta}_{2}=\max \left(  {0,}B_{2}{,B_{3}}\right)  $ for $v_{out}$.
\end{proposition}

We have obtained simple expressions for choosing optimal values of the
regularization parameters. It should be noted that they depend on the
parameter $\beta$ which is unknown. The main advantage of using the parameter
$\beta$ instead of parameters $\lambda_{1}$ and $\lambda_{2}$ or $\eta_{1}$
and $\eta_{2}$ is that $\beta$ is not changed at each iteration of GBM-HRBM
whereas parameters $\lambda_{1}$, $\lambda_{2}$, $\eta_{1}$, $\eta_{2}$ are
defined for each iteration and depend on the corresponding residuals obtained
after the previous iteration. If the number of iterations in GBM-HRBM is $T$,
then the lower value of $\beta$ is defined from the following reasons. The
final prediction of GBM-HRBM is $F_{T}(\mathbf{x})$. Its absolute value is
bounded by absolute values of functions $f_{k}(\mathbf{x})$ and by parameter
$\beta$ as
\begin{equation}
\left \vert F_{T}(\mathbf{x})\right \vert \leq \sum_{k=1}^{T}\left \vert
f_{k}(\mathbf{x})\right \vert \leq T\beta.
\end{equation}

The above implies that $\beta$ is bounded as
\begin{equation}
\beta \geq \max_{\mathbf{x}\in D}\frac{\left \vert F_{T}(\mathbf{x})\right \vert
}{T}%
\end{equation}

The upper bound can be arbitrary, for example, $\beta \leq \max \left \vert
F_{T}(\mathbf{x})\right \vert $.

\section{Algorithms for training and predicting GBM-HRBM}

We provide algorithms for training and predicting GBM-HRBM. First, we consider
algorithms for generating optimal rectangles of two types: closed rectangles
and corners. Second, we show a whole algorithm for implementing GBM-HRBM and
its components.

\subsection{Algorithms for generating rectangles}

The optimal rectangle generation is an essential component of GBM-HRBM.
Therefore, the main idea of Algorithm \ref{alg:random_rect_gen} is to generate
many ($K$) rectangles at each iteration of GBM and to select the best one
which minimizes a cost function $C_{\theta}$. In spite of a general form of
optimization problems (lines 2 and 3 of the algorithm), values $v_{in}$ and
$v_{out}$ (lines 2 and 3 of the algorithm) can be computed by using
(\ref{APR-19}) without solving the optimization problems.

\begin{algorithm}
[H]\caption{Optimal Rectangle Generation}\label{alg:random_rect_gen}

\begin{algorithmic}
[1]\Require Data set $D$, number of generated rectangles $K$

\Ensure Parameters $\theta=(\mathbf{r},v_{in},v_{out})$ of a filled rectangle
$\mathbf{r}$

\Function{Fill}{$r, D$} \Comment{A function of choosing optimal parameters of a rectangle}

\State$v_{in}\leftarrow \arg \min_{v}\sum_{(\mathbf{x},y)\in D}l(y,v)\cdot
\mathbb{I}[\mathbf{x}\in \mathbf{r}]$

\State$v_{out}\leftarrow \arg \min_{v}\sum_{(\mathbf{x},y)\in D}l(y,v)\cdot
\mathbb{I}[\mathbf{x}\notin \mathbf{r}]$

\State \Return$v_{in},v_{out}$

\EndFunction

\For{$i \in {1, \dots, K}$}

\State$\mathbf{r}\leftarrow$\Call{Generate}{D}\Comment{Generating a random rectangle}

\State$v_{in},v_{out}\leftarrow$\Call{Fill}{r, D} \Comment{Choosing optimal values $v_{in},v_{out}$ for the rectangle $\bf{r}$}

\State$\theta \leftarrow(\mathbf{r},v_{in},v_{out})$

\State$C_{\theta}\leftarrow \sum_{(\mathbf{x},y)\in D}\left(  l(y,v_{in}%
)\cdot \mathbb{I}[\mathbf{x}\in \mathbf{r}]+l(y,v_{out})\cdot \mathbb{I}%
[\mathbf{x}\notin \mathbf{r}]\right)  $

\EndFor \State$\theta \leftarrow \arg \min_{\theta}C_{\theta}$

\State \Return$\theta$
\end{algorithmic}
\end{algorithm}

Depending on a type of generated rectangles, we consider two algorithms of the
function \textit{Generate}(). The first one is based on random selection of a
center $\mathbf{c}$ and sizes $w_{1},...,w_{d}$ of each rectangle. Centers of
rectangles are generated from the uniform distribution with parameters defined
by the largest $b^{(j)}$ and smallest $a^{(j)}$ coordinates of all points from
the dataset $D$. In order to cover at least one data point by each rectangle,
we find the nearest point $\alpha$ and farthest point $\beta$ from the $j$-th
coordinate $c^{(j)}$ of the generated center of the rectangle. The width
$\mathbf{w}$ of each rectangle depends on the corresponding distance between
the nearest $\alpha$ (the farthest $\beta$) point and the center. It is shown
as Algorithm \ref{alg:generate_random_rect}. The second algorithm of the
function \textit{Generate}() is based on random selection of a corner (the
rectangle center $\mathbf{c}$) and unbounded elements of vectors $\mathbf{a}$
and $\mathbf{b}$. It is shown as Algorithm \ref{alg:generate_random_corner}.

\begin{algorithm}
[H]\caption{Random Rectangle Generation} \label{alg:generate_random_rect}

\begin{algorithmic}
[1]\Require Data set $D$

\Ensure Generated rectangle $\mathbf{r}$ in the form of closed rectangle

\For{$j \in {1, \dots, d}$}

\State$a^{(j)}\leftarrow \min \left \{  {x_{i}^{(j)}|(\mathbf{x}_{i},y_{i})\in
D}\right \}  $ \Comment{Smallest values of features}

\State$b^{(j)}\leftarrow \max \left \{  {x_{i}^{(j)}|(\mathbf{x}_{i},y_{i})\in
D}\right \}  $ \Comment{Largest values of features}

\State$c^{(j)}\sim \mathcal{U}[a^{(j)},b^{(j)}]$ \Comment{The center of a rectangle is generated from the uniform distribution}

\State$\alpha \leftarrow \min_{i=1,...,n}\left \vert x_{i}^{(j)}-c^{(j)}%
\right \vert $ \Comment{The smallest distance between the center and data points}

\State$\beta \leftarrow \max_{i=1,...,n}\left \vert x_{i}^{(j)}-c^{(j)}%
\right \vert $\Comment{The largest distance between the center and data points}

\State$w^{(j)}\sim \mathcal{U}[\alpha,\beta]$ \Comment{The rectangle width}

\State$r^{(j)}\leftarrow \lbrack c^{(j)}-w^{(j)}/2,c^{(j)}+w^{(j)}/2]$

\EndFor

\State \Return$\mathbf{r}$
\end{algorithmic}
\end{algorithm}

\begin{algorithm}
\caption{Random Corner Generation} \label{alg:generate_random_corner}

\begin{algorithmic}
[1]\Require Data set $D$

\Ensure Generated rectangle in the form of corner $\mathbf{r}$

\For{$j \in {1, \dots, d}$}

\State$a^{(j)}\leftarrow \min \left \{  {x_{i}^{(j)}|(\mathbf{x}_{i},y_{i})\in
D}\right \}  $ \Comment{Smallest values of features}

\State$b^{(j)}\leftarrow \max \left \{  {x_{i}^{(j)}|(\mathbf{x}_{i},y_{i})\in
D}\right \}  $ \Comment{Largest values of features}

\State$c^{(j)}\sim \mathcal{U}[a_{j},b_{j}]$ \Comment{A center of the cornor generated from the uniform distribution}

\State$\zeta \sim \text{Bernoulli}(0.5)$

\If{$\zeta  = 0$}

\State$a^{(j)}\leftarrow-\infty$ and $b^{(j)}\leftarrow c^{(j)}$

\Else \State$a^{(j)}\leftarrow c^{(j)}$ and $b^{(j)}\leftarrow \infty$

\EndIf

\EndFor

\State \Return$\mathbf{r}$
\end{algorithmic}
\end{algorithm}

\subsection{The whole algorithm of GBM-HRBM}

Algorithm \ref{alg:train_hrbm_whole} illustrates GBM with HRBMs. In the
algorithm, we deliberately add rectangles with a zero outside value $v_{out}$
to the resulting set $R$ taking out the sum of all outside values into the
common bias term $q$. This trick allows us to predict faster. The function
\textit{Split}() divides training set $D$ into the training subset $D_{train}$
and the validation subset $D_{val}$. The function \textit{MakeRectangle}()
generates rectangles, but it is slightly different from the similar function
in Algorithm \ref{alg:random_rect_gen}. The difference is that filling of
rectangles (\textit{Fill}()) is performed by using the already calculated
values $g$ and $h$ in (\ref{eq:quardatic_loss_expansion}). The function
\textit{IsValid}() controls whether a new rectangle provides a smaller value
of the loss function on the validation subset.

\begin{algorithm}
[H]%
\caption{The whole training algorithm of GBM-HRBM}\label{alg:train_hrbm_whole}

\begin{algorithmic}
[1]\Require Training data $D$; twice differentiable loss function $l$; number
of iterations for generating rectangles $M$; learning rate $\gamma$; number of
validation attempts $V$

\Ensure The output rectangle set $R$; bias $q$

\Function{IsValid}{$D_{val}, s, \theta$}

\State$C_{val}\leftarrow \sum_{(\mathbf{x}_{i},y_{i})\in D_{val}}l(y_{i}%
,s_{i})$

\State$C_{new}\leftarrow \sum_{(\mathbf{x}_{i},y_{i})\in D_{val}}l(y_{i}%
,s_{i}+\Call{Predict}{\theta, x_i})$

\If{$C_{new} \le C_{val}$} \State \Return{True} \Else \State \Return{False}
\EndIf \EndFunction

\State$q\leftarrow \arg \min_{q}\sum_{(\mathbf{x}_{i},y_{i})\in D}l(y_{i},q)$ \Comment{Initialize the bias term}

\State$\mathbf{s}=(q,\dots,q)^{T}$ \Comment{Initialize the cumulative prediction sum}

\State$(D_{train},D_{val})\leftarrow \Call{Split}{D}$ \Comment{Split dataset into the training and validation subsets}

\State$R\leftarrow \emptyset$ \Comment{Initialize the resulting rectangle set}

\State \Comment{Initial prediction for the $i$-th point is $b$}

\For{$k \in \{ 1, \dots, K\}$} \For{$(x_i, y_i) \in D_{train}$}

\State$\left.  g_{i}\leftarrow \frac{\partial l(y_{i},z)}{\partial
z}\right \vert _{z=s_{i}}$

\State$\left.  h_{i}\leftarrow \frac{\partial^{2}l(y_{i},z)}{\partial z^{2}%
}\right \vert _{z=s_{i}}$

\EndFor

\For{$j \in \{1, \dots, V\}$}

\State$\theta \leftarrow \Call{MakeRectangle}{D_{train}, g, h, l, M}$
\If{\Call{IsValid}{$D_{val}, s, \theta$}} \Comment{If the rectangle passes validation, then add the rectangle}

\State$(\mathbf{r},v_{in},v_{out})\leftarrow \theta$

\State$q\leftarrow q+\gamma \cdot v_{out}$

\For{$(x_i, y_i) \in D$}

\State$\mathbf{s}_{i}\leftarrow \mathbf{s}_{i}+\gamma \cdot \mathbb{I}%
[\mathbf{x}_{i}\in \mathbf{r}]\cdot(v_{in}-v_{out})+\gamma \cdot v_{out}$

\EndFor

\State$R\leftarrow R\cup \left \{  \left(  \mathbf{r},\gamma \cdot(v_{in}%
-v_{out}),0\right)  \right \}  $ \Comment{Addition of the rectangle to the output rectangle set}

\State$(D_{train},D_{val})\leftarrow \Call{Split}{D}$ \Comment{Split the dataset}

\State \textbf{break} \EndIf

\EndFor \EndFor

\State \Return{$R, b$}
\end{algorithmic}
\end{algorithm}

\subsection{Prediction algorithm}

The prediction algorithm, which implements computing the prediction $y$ for a
new example $\mathbf{x}$, is represented as Algorithm
\ref{alg:train_hrbm_standard}.

\begin{algorithm}
[H]\caption{The prediction algorithm}\label{alg:train_hrbm_standard}

\begin{algorithmic}
[1]\Require Data point for prediction $\mathbf{x}$; rectangle set $R$; common
bias $q$

\Ensure Prediction $y$ corresponding to $\mathbf{x}$

\State$y=q$ \Comment{Initialize a cumulative prediction sum}

\For{$\theta \in R$} \State$(\mathbf{r},v_{in},0)\leftarrow \theta$ \Comment{When the ensemble is built, all values $v_{out} = 0$}

\State$y\leftarrow y+\mathbb{I}[\mathbf{x}\in \mathbf{r}]\cdot v_{in}$

\EndFor

\State \Return{$y$}
\end{algorithmic}
\end{algorithm}

\section{Interpretability of GBM-HRBM}

Interpretation means that important features of an analyzed explained example
have to be selected, which significantly impact on the corresponding
prediction provided by a black-box model. By considering interpretation of a
single example, we say about the so-called local interpretation methods
\cite{Guidotti-2019}. They aim to interpret predictions of a black-box model
locally around the considered example. Another (global) interpretation methods
try to interpret predictions taking into account the whole dataset or its
certain part.

One of the most popular post-hoc approaches to interpretation is the
well-known method SHAP \cite{Lundberg-Lee-2017,Strumbel-Kononenko-2010}. SHAP
is widely used in practice and can be viewed as the most promising and
theoretically justified explanation method which fulfils several nice
properties \cite{Lundberg-Lee-2017}. It uses Shapley values
\cite{Shapley-1953} as a concept in coalitional games. According to the
concept, the total gain of a game is distributed among players such that
desirable properties, including efficiency, symmetry, and linearity, dummy are
fulfilled. In the framework of the machine learning, the gain can be viewed as
the machine learning model prediction or the model output, and a player is a
feature of input data. Hence, contributions of features to the model
prediction can be estimated by Shapley values, and the $i$-th feature
importance is defined by the Shapley value denoted as $\phi_{i}$.

Suppose that a prediction $f(\mathbf{x})$ has to be explained for a point
$\mathbf{x}\in \mathbb{R}^{d}$. Denote the index subset of $M=\{1,...,d\}$ as
$S\subseteq M$ and a subset of $k$ features with indices from $S=\{i_{1}%
,...,i_{k}\}$ as $\mathbf{x}^{(S)}=(x^{(i_{1})},...,x^{(i_{k})})$. Let us also
denote $f(\mathbf{x}^{(S)})$ as $\Psi(S)$ for a given $\mathbf{x}$ and $f$.

Shapley values have the following well-known properties:

\textbf{Efficiency}. The total gain is distributed as%
\begin{equation}
\sum_{i=1}^{d}\phi_{k}=\Psi(M)-\Psi(\emptyset).
\end{equation}

\textbf{Symmetry}. If two players with numbers $i$ and $j$ make equal
contributions, i.e., there holds
\begin{equation}
\Psi \left(  S\cup \{i\} \right)  =\Psi \left(  S\cup \{j\} \right)
\end{equation}
for all subsets $S$ which contain neither $i$ nor $j$, then $\phi_{i}=\phi
_{j}$.

\textbf{Linearity}. A linear combination of multiple games $f_{1},...,f_{m}$,
represented as $f(S)=\sum_{k=1}^{m}c_{k}f_{k}(S)$, has gains derived from $f$:
$\phi_{i}(f)=\sum_{k=1}^{m}c_{k}\phi_{i}(f_{k})$ for every $i$.

\textbf{Dummy}. If a player makes zero contribution, i.e., $\Psi \left(
S\cup \{j\} \right)  =\Psi \left(  S\right)  $ for a player $j$ and all
$S\subseteq M\backslash \{j\}$, then $\phi_{j}=0$.

If all the above properties are satisfied, then we can write the following
expression for computing Shapley values:%
\begin{equation}
\phi_{i}=\sum_{S\subseteq M\backslash \{i\}}\frac{\#S!\left(  d-\#S-1\right)
!}{d!}\left[  \Psi \left(  S\cup \{i\} \right)  -\Psi \left(  S\right)  \right]
, \label{SHAP_1}%
\end{equation}
where $\#S$ is the number of elements in $S$.

Let us define the method for calculating the feature subset function $\Psi(S)$
for HRBM as follows:
\begin{equation}
\Psi(S)=\mathbb{I}[\mathbf{x}^{(S)}\in \mathbf{r}^{(S)}]\cdot(v_{in}%
-v_{out})+v_{out}, \label{SHAP_2}%
\end{equation}
where $\mathbf{r}^{(S)}$ is a rectangle that includes only features with
indices from subset $S$.

For the empty set of features, it is natural to set the value of the indicator
function equal to 1. Then there holds:
\begin{equation}
\Psi(\emptyset)=(v_{in}-v_{out})+v_{out}=v_{in}. \label{phi0}%
\end{equation}

\begin{proposition}
\label{prop:rect_shap_1}The contribution of the $i$-th feature of HRBM is
determined as follows:%
\begin{equation}
\phi_{i}=%
\begin{cases}
0, & x^{(i)}\in r^{(j)},\\
\dfrac{v_{out}-v_{in}}{\sum_{j=1}^{d}\mathbb{I}[x^{(i)}\notin r^{(j)}]}, &
x^{(i)}\notin r^{(j)}.
\end{cases}
\label{SHAP_3}%
\end{equation}

\end{proposition}

The main advantage of this approach is its isolation from the training data
set: the calculated contributions depend solely on the structure of the model
(rectangles and their inside values), but do not depend on the data, allowing
us to explore with their help separately the behavior of the model. That is
why we call the method as the \emph{model-based SHAP}. On the other hand, this
approach does not take into account how the model predictions are supported by
the training data, which makes it difficult to extract useful information
about the task. Therefore, we consider also an original definition
\cite{Lundberg-Lee-2017} of the function of a subset of features through
expectations:
\begin{equation}
\tilde{\Psi}(S)=\mathbb{E}[f(X)\mid X_{S}=\mathbf{x}^{(S)}],
\label{SHAP-exp-est}%
\end{equation}%
\begin{equation}
\tilde{\Psi}(\emptyset)=\mathbb{E}[f(X)].
\end{equation}

This method is called the \emph{data-based SHAP}.

\begin{proposition}
\label{prop:rect_shap_2}By using the original definition of Shapley values
through expectations \cite{Lundberg-Lee-2017}, the contribution $\phi_{i}$ of
the $i$-th feature of HRBM is determined as follows:

For each $i$-th feature such that $x^{(i)}\notin r^{(i)}$, there holds
$\phi_{i}=\tilde{\phi}$, otherwise the contribution $\phi_{i}$ is computed as
\begin{equation}
\phi_{i}=\sum_{S\subset \{1,\dots,d\} \setminus \{i\}}\frac{\left \vert
S\right \vert !(d-\left \vert S\right \vert -1)!}{d!}(\tilde{\Psi}(S\cup
\{i\})-\tilde{\Psi}(S)), \label{SHAP_6}%
\end{equation}
where
\begin{align}
\tilde{\Psi}(S\cup \{i\})-\tilde{\Psi}(S)  &  =\frac{v_{in}-v_{out}}{N}%
\cdot \mathbb{I}[\mathbf{x}^{(S)}\in \mathbf{r}^{(S)}]\nonumber \\
\times \sum_{t=1}^{N}\mathbb{I}[\mathbf{x}_{t}^{(\overline{S}\setminus \{i\})}
&  \in \mathbf{r}^{(\overline{S}\setminus \{i\})}]\cdot \mathbb{I}[x_{t}%
^{(i)}\notin r^{(i)}]. \label{SHAP_7}%
\end{align}

Here $\overline{S}=M\setminus S$ is the set of features which are not included
in $S$. The sum in (\ref{SHAP_6}) is calculated only for subsets $S$ which
fulfil condition $\mathbf{x}^{(S)}\in \mathbf{r}^{(S)}$.

Value $\tilde{\phi}$ is calculated from the efficiency property as
\begin{equation}
\tilde{\phi}=\frac{\tilde{\Psi}(\{1,\dots,d\})-\tilde{\Psi}(\emptyset
)-\sum_{i=1}^{d}\phi_{i}\cdot \mathbb{I}[x^{(i)}\in r^{(i)}]}{\sum_{j=1}%
^{d}\mathbb{I}[x^{(j)}\notin r^{(j)}]}. \label{SHAP_8}%
\end{equation}

\end{proposition}

Thus, we need first to find values for all features for which the explained
example is inside the rectangle using (\ref{SHAP_6}), and then using
(\ref{SHAP_8}) to calculate Shapley values for the remaining features. In
order to obtain Shapley values for an arbitrary HRBM ensemble, for example,
boosting, we use the third property (linearity) of Shapley values. According
to this property, the contribution of each feature of the weighted ensemble is
represented as the sum of the contributions of each model with the appropriate weights.

\begin{remark}
It is important to point out that the considered GBM-HRBM is not a black-box
model because, in addition to the input data and predictions of the model, we
use information about generated rectangles at iterations of the boosting. On
the other hand, GBM-HRBM can be used as an interpretable meta-model
approximating some black-box model, for instance, a deep neural network.
Moreover, GBM-HRBM can play a double role. First, it can be regarded as an
interpretable approximating model like the linear model. In this case,
GBM-HRBM is trained on the feature vectors and the corresponding predicted
values provided by the black-box model. As a result, GBM-HRBM extends the set
of interpretable models. Second, GBM-HRBM can be viewed as a student model for
the teacher black-box model in the knowledge distillation framework. In
addition, GBM-HRBM can be the interpretable meta-model approximating the
black-box model and a student model in distillation similarly to the DeepVID
model \cite{Wang-Gou-etal-2019}.
\end{remark}

\section{Numerical experiments}

In order to study the proposed GBM-HRBM for solving regression problems, we
apply datasets which are taken from open sources, in particular: Diabetes can
be found in the corresponding R Packages; Friedman 1, 2 3 are described at
site: https://www.stat.berkeley.edu/\symbol{126}breiman/bagging.pdf;
Scikit-Learn Sparse Uncorrelated (Sparse) datasets are available in package
\textquotedblleft Scikit-Learn\textquotedblright. The proposed algorithm is
evaluated and investigated also by the following publicly available datasets
from the UCI Machine Learning Repository \cite{Dua:2019} (short notations are
given in brackets): Auto MPG, Boston Housing (Boston), Concrete, Forest Fires,
Yacht Hydrodynamics (Yacht), Airfoil. We also use datasets Chscase\_Census2
(CCC2), ERA, FruitFly, Fish Catch, LiverDisorders, MachineCPU from OpenML
https://www.openml.org. A brief introduction about these data sets are given
in Table \ref{t:regres_datasets} where $d$ and $N$ are numbers of features and
examples, respectively. A more detailed information can be found from the
aforementioned data resources.%

\begin{table}[tbp] \centering
\caption{A brief introduction about datasets for regression}%
\begin{tabular}
[c]{cccccc}\hline
Dataset & $d$ & $N$ & Dataset & $d$ & $N$\\ \hline
Airfoil & $5$ & $1503$ & Friedman1 & $10$ & $100$\\
AutoMpg & $8$ & $398$ & Friedman2 & $4$ & $100$\\
Boston & $13$ & $506$ & Friedman3 & $4$ & $100$\\
CCC2 & $8$ & $400$ & FruitFly & $5$ & $125$\\
Concrete & $8$ & $1030$ & LiverDisorders & $6$ & $345$\\
Diabetes & $10$ & $442$ & MachineCPU & $7$ & $209$\\
ERA & $4$ & $1000$ & Sparse & $10$ & $100$\\
FishCatch & $7$ & $159$ & Yacht & $6$ & $308$\\
ForestFire & $13$ & $517$ &  &  & \\ \hline
\end{tabular}
\label{t:regres_datasets}%
\end{table}%

We use the coefficient of determination denoted $R^{2}$ for the regression
evaluation. The greater the value of the coefficient of determination, the
better results we get. The best results in all tables are shown in bold.
$F_{1}$ score is used in the classification experiments as an accuracy measure
which takes into account the possible class imbalance.

The proposed GBM-HRBM model is compared with ensemble-based models including
RFs, ERTs, and GBM with decision trees as base models. Numbers of trees $100$,
$200$, $300$, $400$, $500$ in RFs and ERTs are tested, choosing those leading
to the best results.

In order to optimize the model parameters in numerical experiments, we perform
a 5-fold cross-validation on the training set which consists of different
numbers of randomly selected examples. The cross-validation is performed with
$50$ repetitions. This procedure is realized by considering all possible
values of the regularization parameter $\beta$ and other tuning parameters in
a predefined grid. Their values are also tested, choosing those leading to the
best results.

The code implementing GBM-HRBM can be found at: https://github.com/andruekonst/HRBM.

\subsection{Regression}

In order to compare GBM-HRBM with other ensemble-based models, measures
$R^{2}$ for the RF, ERT, GBM with decision trees as base models, and for
GBM-HRBM are computed for several real datasets and shown in Table
\ref{t:regr_rect_1}. \ Measures $R^{2}$ for GBM-HRBM are obtained as the best
values among two types of HRBMs: rectangles and corners. It can be seen from
Table \ref{t:regr_rect_1} that GBM-HRBM outperforms the aforementioned
ensemble-based models for 13 from 17 datasets. Moreover, the difference
between values of $R^{2}$ is large for datasets Sparse ($0.124$), Diabetes
($0.046$), Fridman1 ($0.041$). At the same time, there are datasets (Airfoil,
Concrete, Friedman2, Yacht) for which one of the models (RF, ERT, GBM) provide
better results in comparison with GBM-HRBM.

Let us compare values of $R^{2}$ obtained for GBM-HRBM with the best results
provided by one of the models: RF, ERT, GBM. For comparison, we can apply the
$t$-test. According to \cite{Demsar-2006}, the $t$-statistics is distributed
in accordance with the Student distribution with $17-1$ degrees of freedom
($17$ datasets). The obtained p-value is $p=0.045$. We can conclude that the
outperformance of GBM-HRBM is statistically significant because $p<0.05$.%

\begin{table}[tbp] \centering
\caption{$R^2$ measures for comparison of the RF, ERT, GBM, GBM-HRBM in the regression tasks}
\begin{tabular}
[c]{ccccc}\hline
Dataset & RF & ERT & GBM & GBM-HRBM\\ \hline
Airfoil & $0.909$ & $0.910$ & $\mathbf{0.952}$ & $0.922$\\
AutoMpg & $0.870$ & $0.871$ & $0.863$ & $\mathbf{0.882}$\\
Boston & $0.859$ & $0.863$ & $0.861$ & $\mathbf{0.876}$\\
CCC2 & $-0.048$ & $-0.030$ & $-0.022$ & $\mathbf{-0.018}$\\
Concrete & $0.898$ & $0.901$ & $\mathbf{0.933}$ & $0.927$\\
Diabetes & $0.426$ & $0.431$ & $0.422$ & $\mathbf{0.477}$\\
ERA & $0.361$ & $0.360$ & $0.361$ & $\mathbf{0.364}$\\
FishCatch & $0.951$ & $0.957$ & $0.948$ & $\mathbf{0.974}$\\
ForestFire & $-0.014$ & $-0.013$ & $-0.012$ & $\mathbf{-0.009}$\\
Friedman1 & $0.778$ & $0.803$ & $0.880$ & $\mathbf{0.921}$\\
Friedman2 & $0.983$ & $\mathbf{0.994}$ & $0.986$ & $0.991$\\
Friedman3 & $0.874$ & $0.921$ & $0.908$ & $\mathbf{0.923}$\\
FruitFly & $-0.102$ & $-0.099$ & $-0.087$ & $\mathbf{-0.051}$\\
LiverDisorders & $0.160$ & $0.162$ & $0.126$ & $\mathbf{0.201}$\\
MachineCPU & $0.864$ & $0.856$ & $0.827$ & $\mathbf{0.872}$\\
Sparse & $0.627$ & $0.682$ & $0.681$ & $\mathbf{0.811}$\\
Yacht & $0.995$ & $0.995$ & $\mathbf{0.998}$ & $0.996$\\ \hline
\end{tabular}
\label{t:regr_rect_1}%
\end{table}%

Another interesting question is how different types of the regularization
impact on the accuracy of GBM-HRBM. We compare three types of regularization,
including the $L_{2}$-norm of step height penalty (see (\ref{APR-48})), the
$L_{1}$- and $L_{2}$-norms of the standard regularization (see (\ref{APR-40}%
)). The corresponding values of $R^{2}$ are shown in Table \ref{t:regr_rect_2}%
. One can see from Table \ref{t:regr_rect_2} that it is difficult to select
the best type of regularization. Each type demonstrates outperforming results
for several datasets. Therefore, it makes sense to analyze all types of
regularization for new datasets.%

\begin{table}[tbp] \centering
\caption{$R^2$ measures for comparison of different types of regularization for GBM-HRBM}
\begin{tabular}
[c]{cccc}\hline
Dataset & Step height penalty $L_{2}$ & Standard $L_{1}$ & Standard $L_{2}%
$\\ \hline
Airfoil & $0.879$ & $0.873$ & $\mathbf{0.922}$\\
AutoMpg & $0.881$ & $\mathbf{0.882}$ & $0.881$\\
Boston & $\mathbf{0.876}$ & $0.866$ & $0.873$\\
CCC2 & $\mathbf{-0.018}$ & $\mathbf{-0.018}$ & $\mathbf{-0.018}$\\
Concrete & $0.925$ & $0.911$ & $\mathbf{0.927}$\\
Diabetes & $\mathbf{0.477}$ & $0.473$ & $\mathbf{0.477}$\\
ERA & $\mathbf{0.364}$ & $\mathbf{0.364}$ & $0.363$\\
FishCatch & $\mathbf{0.974}$ & $\mathbf{0.974}$ & $\mathbf{0.974}$\\
ForestFire & $-0.010$ & $-0.010$ & $\mathbf{-0.009}$\\
Friedman1 & $\mathbf{0.921}$ & $0.906$ & $0.920$\\
Friedman2 & $\mathbf{0.991}$ & $0.982$ & $0.990$\\
Friedman3 & $0.917$ & $0.885$ & $\mathbf{0.923}$\\
FruitFly & $\mathbf{-0.051}$ & $\mathbf{-0.051}$ & $-0.052$\\
LiverDisorders & $0.188$ & $0.188$ & $\mathbf{0.201}$\\
MachineCPU & $0.870$ & $0.869$ & $\mathbf{0.872}$\\
Sparse & $0.808$ & $0.773$ & $\mathbf{0.811}$\\
Yacht & $0.995$ & $0.989$ & $\mathbf{0.996}$\\ \hline
\end{tabular}
\label{t:regr_rect_2}%
\end{table}%
%

\begin{figure}
[ptb]
\begin{center}
\includegraphics[
height=3.634in,
width=3.7105in
]%
{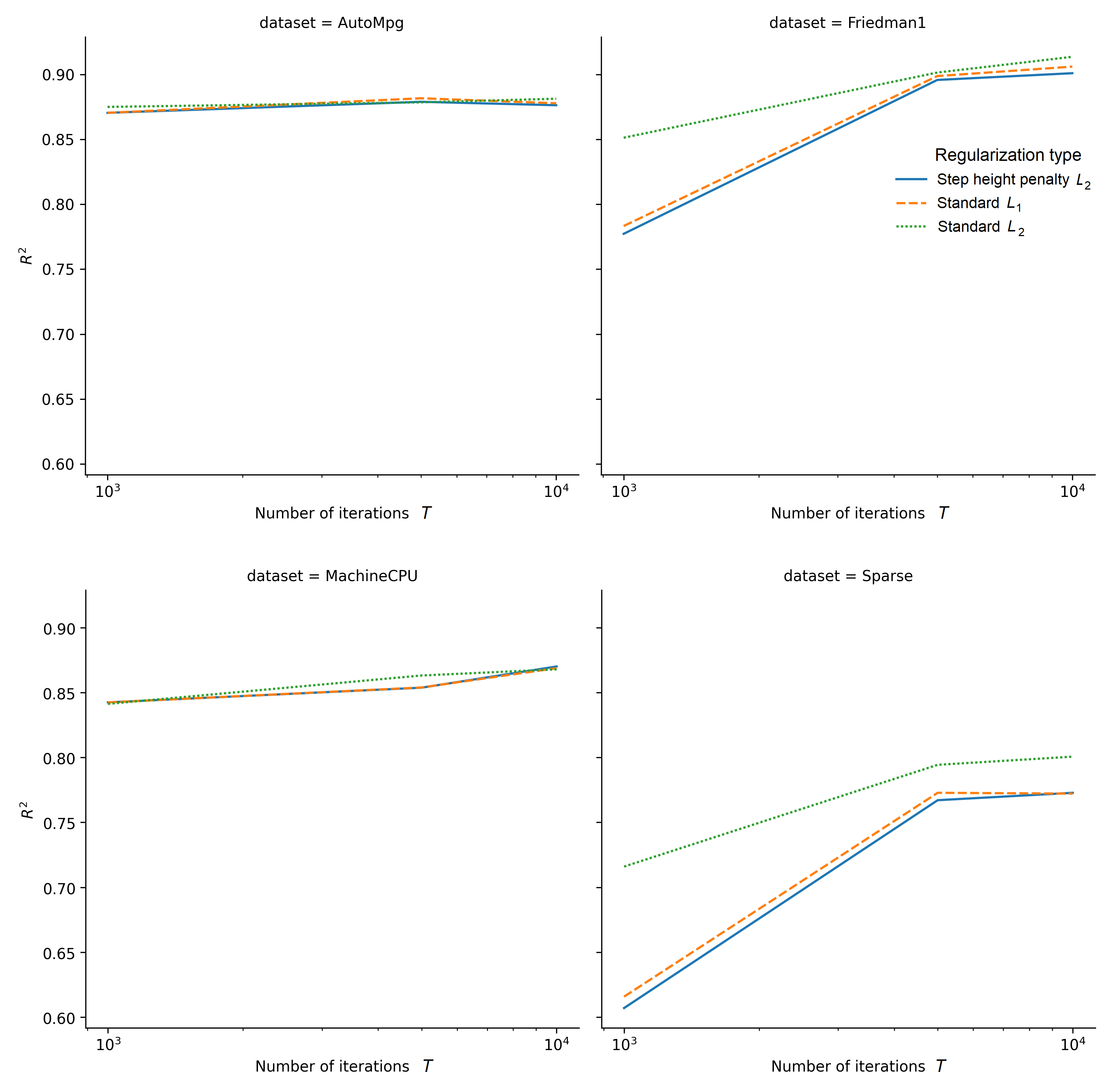}%
\caption{$R^{2}$ as a function of the number of the GBM-HRBM iterations for
four datasets by different regularization types}%
\label{f:reg_n_hrbm_reg}%
\end{center}
\end{figure}

Fig. \ref{f:reg_n_hrbm_reg} shows how the model accuracy depends on the number
of iterations $T$ for four datasets (AutoMpg, Friedman1, MachineCPU, Sparse)
under condition of using three regularization types (step height penalty using
Proposition \ref{prop:regular2} and standard $L_{1}$, $L_{2}$ regularizations
using Proposition \ref{prop:regular1}). The step height penalty and standard
$L_{1}$, $L_{2}$ regularizations are depicted by solid, dashed and dotted
lines. One can see from Fig. \ref{f:reg_n_hrbm_reg} that $R^{2}$ increases
with the number of iterations. However, after some number of iterations, the
accuracy almost does not increase. It can also be seen from Fig.
\ref{f:reg_n_hrbm_reg} that the regularization type does not significantly
impact on the accuracy except for the case of Friedman1 and Sparse datasets
where the standard $L_{2}$ regularization provides better results.%

\begin{figure}
[ptb]
\begin{center}
\includegraphics[
height=4.0077in,
width=4.0438in
]%
{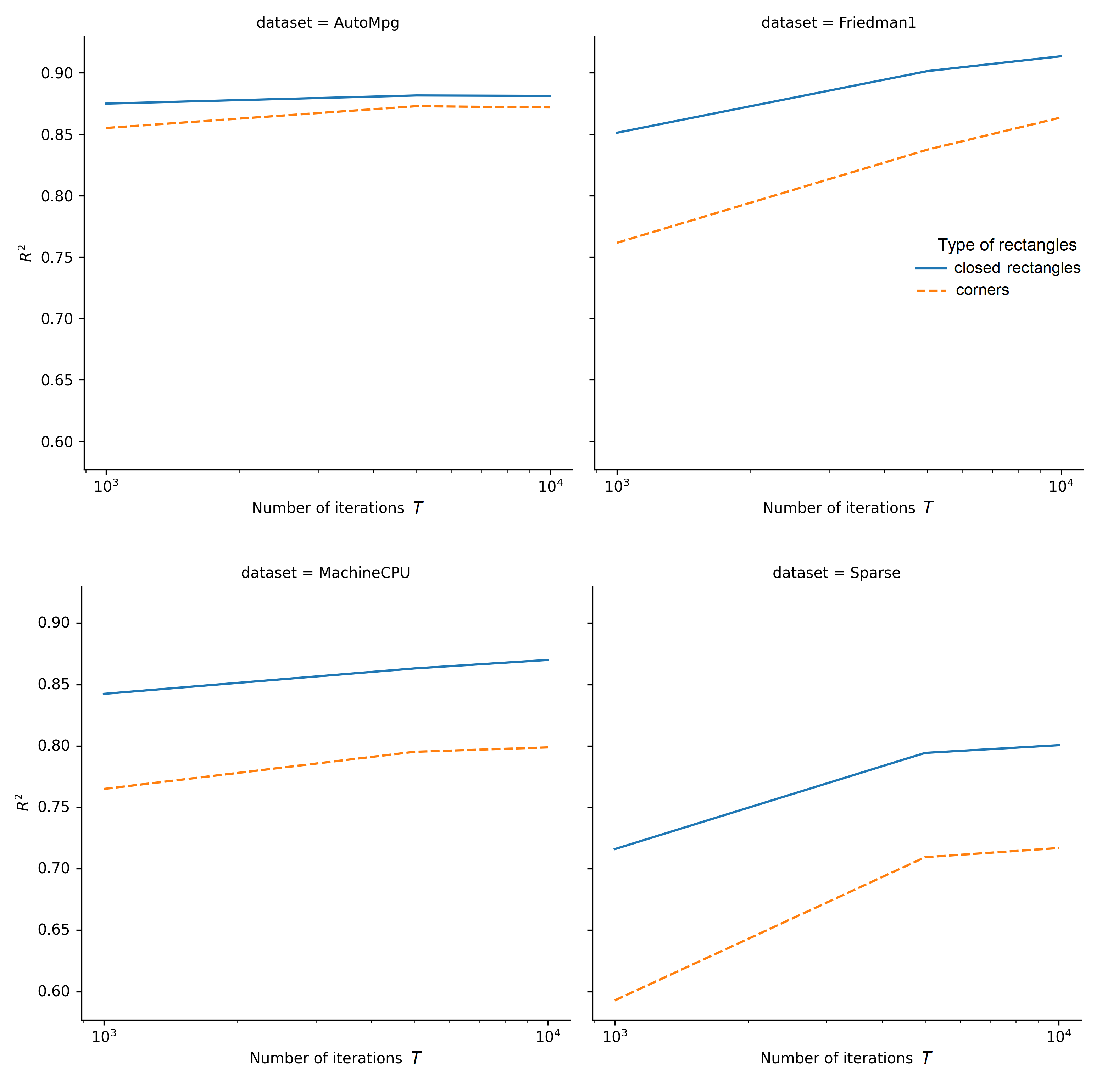}%
\caption{$R^{2}$ as a function of the number of the GBM-HRBM iterations for
four datasets by different types of rectangles}%
\label{f:reg_n_hrbm_kind}%
\end{center}
\end{figure}

The same dependencies of $R^{2}$ on numbers of iterations are shown in Fig.
\ref{f:reg_n_hrbm_kind}, but, in contrast to the previous experiment, we study
how the type of rectangles impacts on the accuracy. Functions under conditions
of using corners and closed rectangles are depicted by the solid and dashed
lines respectively. It can be seen from Fig. \ref{f:reg_n_hrbm_kind} that the
tendency of functions does not differ from the same functions in Fig.
\ref{f:reg_n_hrbm_reg}. However, the model with corners outperforms the model
with closed rectangles.%

\begin{figure}
[ptb]
\begin{center}
\includegraphics[
height=3.758in,
width=3.9339in
]%
{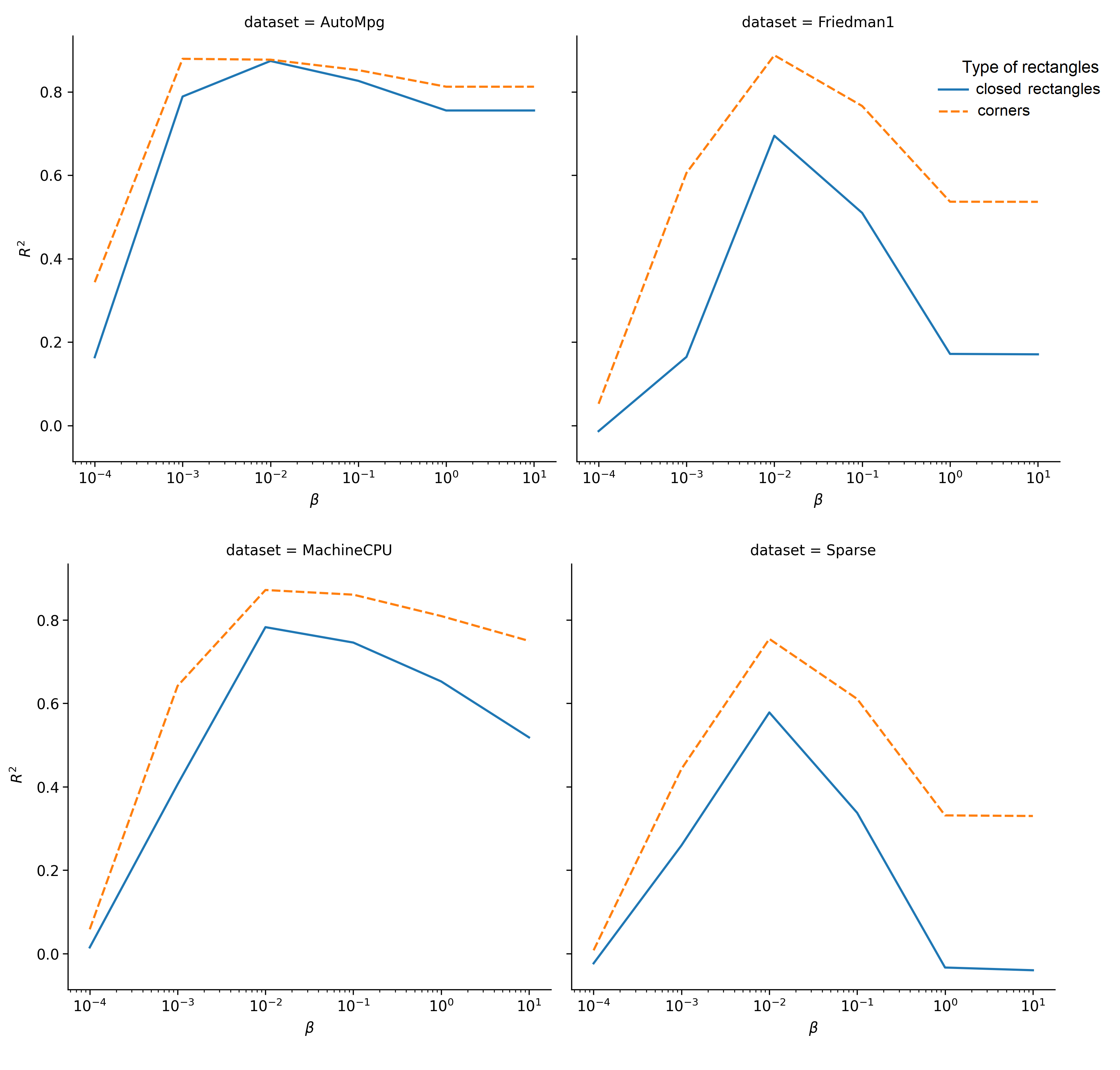}%
\caption{$R^{2}$ as a function of the regularization parameter $\beta$ for
four datasets (AutoMpg, Friedman1, MachineCPU, Sparse) by using corners and
closed rectangles }%
\label{f:reg_l2_beta_kind}%
\end{center}
\end{figure}

The next question is how the $R^{2}$ measure depends on the regularization
parameter $\beta$. We consider the standard $L_{2}$ regularization
(\ref{APR-40}). The corresponding dependencies are depicted in Fig.
\ref{f:reg_l2_beta_kind} where GBM-HRBM with the $L_{2}$ regularization is
trained on the same four regression datasets (AutoMpg, Friedman1, MachineCPU,
Sparse) with a fixed number of iterations of GBM-HRBM. Moreover, two HRBMs are
studied for every dataset: corners and closed rectangles depicted by dashed
and solid lines, respectively. It can be seen from Fig.
\ref{f:reg_l2_beta_kind} that there exists an optimal value of $\beta$ for all
datasets. This peculiarity is very important because we do not need to tune
the parameter $\lambda_{2}$ at each iteration of GBM-HRBM. We tune only
$\beta$ and compute different $\lambda_{2}$ at each iteration by using
Proposition \ref{prop:regular3}. Another interesting observation from Fig.
\ref{f:reg_l2_beta_kind} is that GBM-HRBM with corners shows better results in
comparison with GBM-HRBM with closed rectangles almost for all $\beta$. In
order to confirm this observation, we consider corners and closed rectangles
for other datasets. The corresponding values of $R^{2}$ are shown in Table
\ref{t:regr_rect_3}. It is seen from the results that corners provide
outperforming results for most datasets. If we again apply the $t$-test to
these results, then we obtain p-value equal to $p=0.004$. It is obvious that
corners give the statistically significant outperformance.%

\begin{table}[tbp] \centering
\caption{$R^2$ measures for comparison of different types of rectangles: corners and closed rectangles}%
\begin{tabular}
[c]{ccc}\hline
Dataset & Corners & Rectangles\\ \hline
Airfoil & $\mathbf{0.922}$ & $0.891$\\
AutoMpg & $\mathbf{0.882}$ & $0.877$\\
Boston & $\mathbf{0.876}$ & $0.857$\\
CCC2 & $\mathbf{-0.018}$ & $\mathbf{-0.018}$\\
Concrete & $\mathbf{0.927}$ & $0.922$\\
Diabetes & $\mathbf{0.477}$ & $0.460$\\
ERA & $\mathbf{0.364}$ & $0.356$\\
FishCatch & $\mathbf{0.974}$ & $0.957$\\
ForestFire & $-0.010$ & $\mathbf{-0.009}$\\
Friedman1 & $\mathbf{0.921}$ & $0.879$\\
Friedman2 & $\mathbf{0.991}$ & $0.962$\\
Friedman3 & $\mathbf{0.923}$ & $0.906$\\
FruitFly & $\mathbf{-0.051}$ & $-0.053$\\
LiverDisorders & $0.188$ & $\mathbf{0.201}$\\
MachineCPU & $\mathbf{0.872}$ & $0.821$\\
Sparse & $\mathbf{0.811}$ & $0.730$\\
Yacht & $\mathbf{0.996}$ & $0.987$\\ \hline
\end{tabular}
\label{t:regr_rect_3}%
\end{table}%

\begin{figure}
[ptb]
\begin{center}
\includegraphics[
height=3.67in,
width=3.7553in
]%
{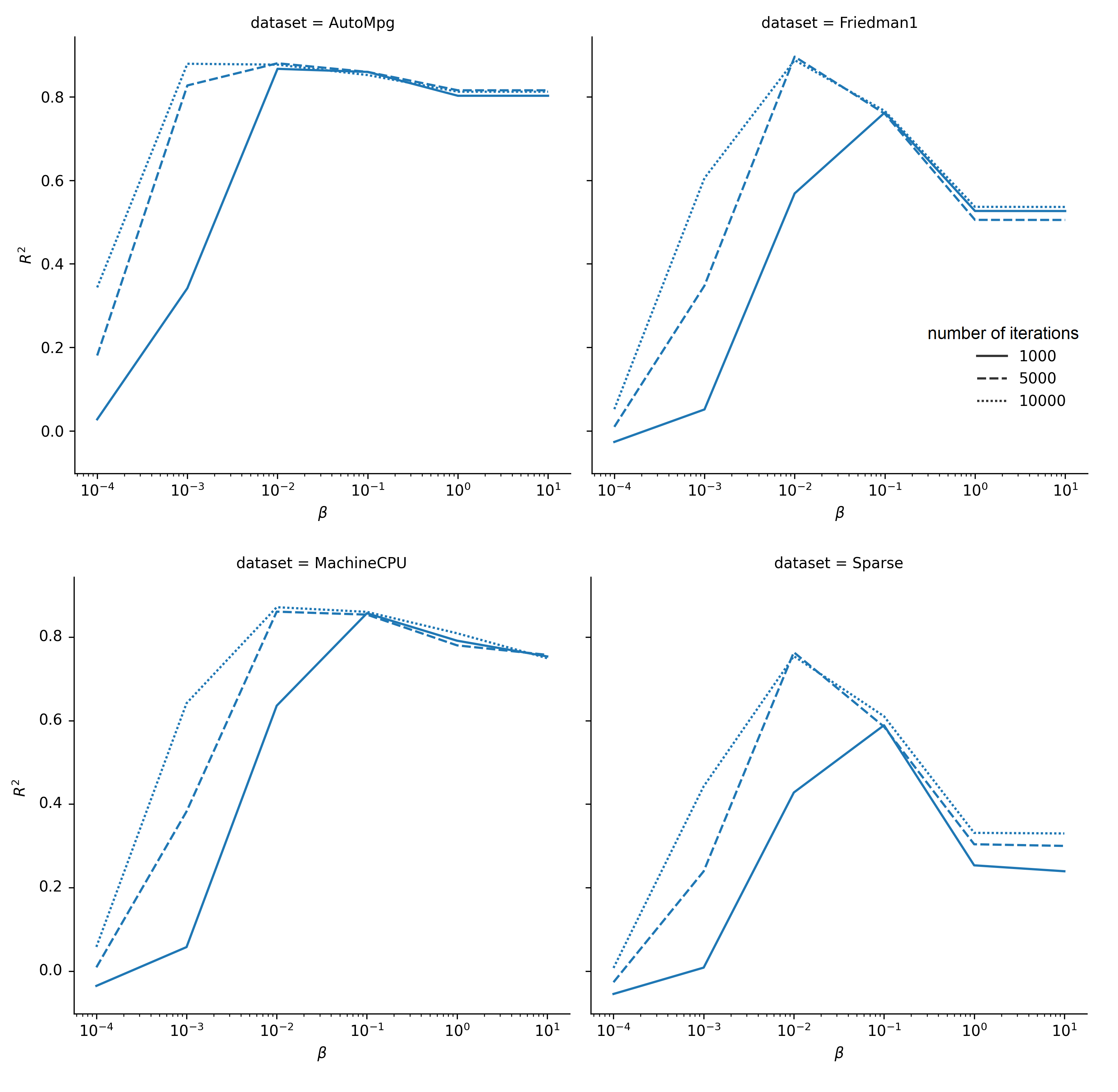}%
\caption{$R^{2}$ as a function of the regularization parameter $\beta$ for
four datasets by different numbers $T=1000$, $5000$, $10000$ of iterations}%
\label{f:reg_l2_beta_n}%
\end{center}
\end{figure}

Fig. \ref{f:reg_l2_beta_n} illustrates how the accuracy $R^{2}$ depends on the
regularization parameter $\beta$ by using the standard $L_{2}$ regularization.
GBM-HRBM is trained on the same four regression datasets with different
numbers $T=1000$, $5000$, $10000$ of iterations, depicted by solid, dashed and
dotted lines, respectively. We again see from Fig. \ref{f:reg_l2_beta_n} that
there exist optimal values of $\beta$ corresponding to the largest accuracy
measure. It is interesting to observe from Fig. \ref{f:reg_l2_beta_n} that
optimal values of $\beta$ almost coincide when the number of iterations is
larger than 1000. This implies that the regularization does not impact on the
prediction accuracy after some number of iterations.%

\subsection{Classification}

To study the proposed GBM-HRBM for solving classification problems, we apply
datasets which are taken from the UCI Machine Learning Repository
\cite{Dua:2019}, in particular, Balance Scale (Balance), Car Evaluation (Car),
Dermatology, Diabetic Retinopathy (Retinopathy), Glass Identification (Glass)
Haberman's Survival (Haberman), Ionosphere, Seeds, Seismic-Bumps (Seismic),
Soybean, Teaching Assistant Evaluation (Teaching), Tic-Tac-Toe Endgame (TTT),
Website Phishing (Website), Wholesale Customer (Wholesale). Short notations of
datasets are given in brackets. The dataset Diabetes is taken from OpenML at
https://www.openml.org/. Table \ref{t:class_datasets} shows the number of
features $d$ for the corresponding data set, the number of examples $N$, and
the number of classes $C$. More detailed information can be found from the
data resources. Parameters of experiments coincide with similar parameters in
experiments with the regression models.%

\begin{table}[tbp] \centering
\caption{A brief introduction about the classification data sets}%
\begin{tabular}
[c]{cccccccc}\hline
Dataset & $d$ & $N$ & $C$ & Dataset & $d$ & $N$ & $C$\\ \hline
Balance & $4$ & $625$ & $3$ & Seeds & $7$ & $210$ & $3$\\
Car & $6$ & $1728$ & $4$ & Seismic & $18$ & $2584$ & $2$\\
Dermatology & $33$ & $366$ & $6$ & Soybean & $35$ & $47$ & $4$\\
Diabetes & $9$ & $768$ & $2$ & TTT & $27$ & $957$ & $2$\\
Glass & $10$ & $214$ & $6$ & Teaching & $5$ & $151$ & $3$\\
Haberman & $3$ & $306$ & $2$ & Website & $9$ & $1353$ & $3$\\
Ionosphere & $34$ & $351$ & $2$ & Wholesale & $6$ & $440$ & $3$\\
Retinopathy & $20$ & $1151$ & $2$ &  &  &  & \\ \hline
\end{tabular}
\label{t:class_datasets}%
\end{table}%

First, we compare values of $F_{1}$ obtained for GBM-HRBM with the best
results provided by one of the models: RF, ERT, GBM. Results are shown in
Table \ref{t:class_rect_1}. The largest differences between values of $F_{1}$
are for datasets Balance ($\allowbreak0.168\,$), Dermatology ($0.112$),
Seismic ($\allowbreak0.071\,$). For comparison, we again apply the $t$-test.
The obtained p-value is $p=0.0066$. This implies that the outperformance of
GBM-HRBM is statistically significant.%

\begin{table}[tbp] \centering
\caption{$F1$ measures for comparison of different types of the RF, ERT, GBM, GBM-HRBM in the classification tasks}%
\begin{tabular}
[c]{ccccc}\hline
Dataset & RF & ERT & GBM & GBM-HRBM\\ \hline
Balance & $0.599$ & $0.614$ & $0.662$ & $\mathbf{0.830}$\\
Car & $0.837$ & $0.873$ & $\mathbf{0.985}$ & $0.968$\\
Dermatology & $0.201$ & $0.191$ & $0.196$ & $\mathbf{0.313}$\\
Diabetes & $0.722$ & $0.703$ & $0.719$ & $\mathbf{0.729}$\\
Glass & $0.683$ & $0.602$ & $0.658$ & $\mathbf{0.697}$\\
Haberman & $0.514$ & $0.439$ & $0.487$ & $\mathbf{0.594}$\\
Ionosphere & $0.926$ & $\mathbf{0.934}$ & $0.922$ & $0.932$\\
Retinopathy & $0.677$ & $0.690$ & $0.694$ & $\mathbf{0.719}$\\
Seeds & $0.928$ & $0.929$ & $0.929$ & $\mathbf{0.940}$\\
Seismic & $0.484$ & $0.483$ & $0.490$ & $\mathbf{0.561}$\\
Soybean & $0.989$ & $0.990$ & $0.961$ & $\mathbf{1.000}$\\
TTT & $0.968$ & $0.978$ & $\mathbf{0.998}$ & $0.996$\\
Teaching & $0.580$ & $0.546$ & $0.575$ & $\mathbf{0.633}$\\
Website & $0.792$ & $0.749$ & $0.841$ & $\mathbf{0.868}$\\
Wholesale & $0.278$ & $0.278$ & $0.278$ & $\mathbf{0.342}$\\ \hline
\end{tabular}
\label{t:class_rect_1}%
\end{table}%

In order to compare corners and rectangles as base models, we compute values
of $F_{1}$ score for the classification datasets taking GBM-HRBM with corners
and rectangles. The corresponding results are shown in Table
\ref{t:class_rect_2}. One can again see that corners provide outperforming
results for most datasets. The application of the $t$-test shows that p-value
in this case is equal to $p=0.031$. It implies that corners again give the
statistically significant outperformance.%

\begin{table}[tbp] \centering
\caption{$F1$ measures for comparison of classifiers with different types of rectangles: corners and closed rectangles}%
\begin{tabular}
[c]{ccc}\hline
Dataset & Corners & Rectangles\\ \hline
Balance & $\mathbf{0.830}$ & $0.803$\\
Car & $\mathbf{0.968}$ & $0.955$\\
Dermatology & $\mathbf{0.313}$ & $0.185$\\
Diabetes & $\mathbf{0.729}$ & $0.725$\\
Glass & $\mathbf{0.697}$ & $0.692$\\
Haberman & $0.591$ & $\mathbf{0.594}$\\
Ionosphere & $\mathbf{0.932}$ & $0.915$\\
Retinopathy & $\mathbf{0.719}$ & $0.689$\\
Seeds & $\mathbf{0.940}$ & $0.938$\\
Seismic & $\mathbf{0.561}$ & $0.554$\\
Soybean & $\mathbf{1.000}$ & $\mathbf{1.000}$\\
TTT & $\mathbf{0.996}$ & $0.962$\\
Teaching & $\mathbf{0.633}$ & $0.624$\\
Website & $\mathbf{0.868}$ & $0.863$\\
Wholesale & $\mathbf{0.342}$ & $0.323$\\ \hline
\end{tabular}
\label{t:class_rect_2}%
\end{table}%

\subsection{SHAP and GBM-HRBM}

\begin{figure}
[ptb]
\begin{center}
\includegraphics[
height=2.4898in,
width=4.9614in
]%
{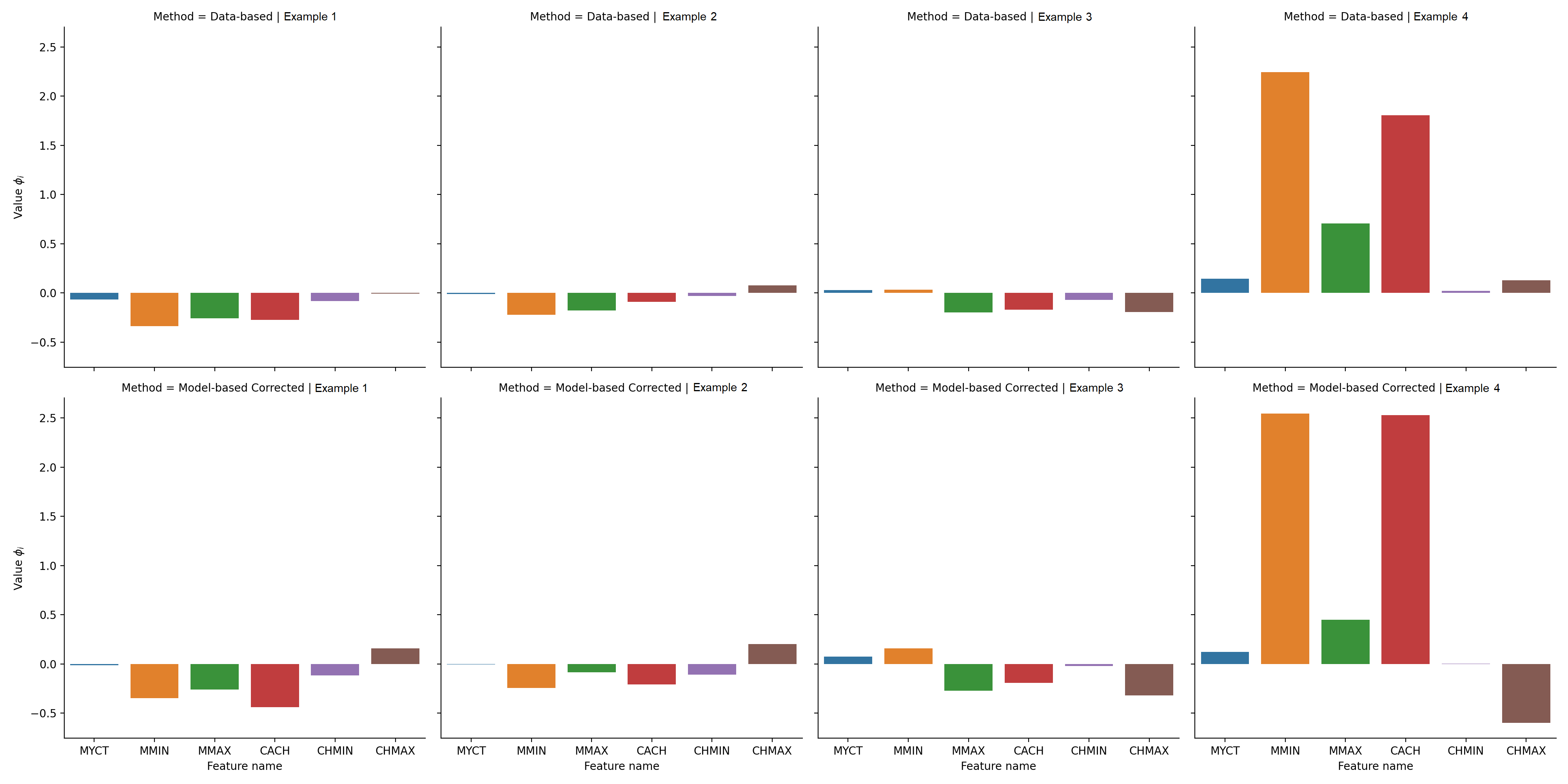}%
\caption{Comparison of Shapley values obtained by using the data-based SHAP
(the first row of pictures) and the model-based SHAP (the second row of
pictures) for the dataset MachineCPU}%
\label{f:SHAP1}%
\end{center}
\end{figure}

In order to study modifications of SHAP for GBM-HRBM, we use the regression
dataset MachineCPU. Fig. \ref{f:SHAP1} shows Shapley values $\phi_{i}$ of all
features (MYCT, MMIN, MMAX, CACH, CHMIN, CHMAX) obtained by two methods: the
data-based SHAP and the model-based SHAP. Four randomly selected examples from
the dataset are used for comparison of methods. We do not provide Shapley
values obtained by using the original SHAP because they totally coincide with
the corresponding values provided by the data-based method. It follows from
Fig. \ref{f:SHAP1} that most Shapley values calculated by using the data-based
SHAP are similar to the corresponding Shapley values calculated by using the
model-based SHAP. It should be noted that there is some divergence of results.
In particular, Shapley values of the feature CHMAX in the example 4 are quite
different and even have different signs. However, this case can be regarded as
an exception to the rule.%

\begin{figure}
[ptb]
\begin{center}
\includegraphics[
height=2.5496in,
width=3.3939in
]%
{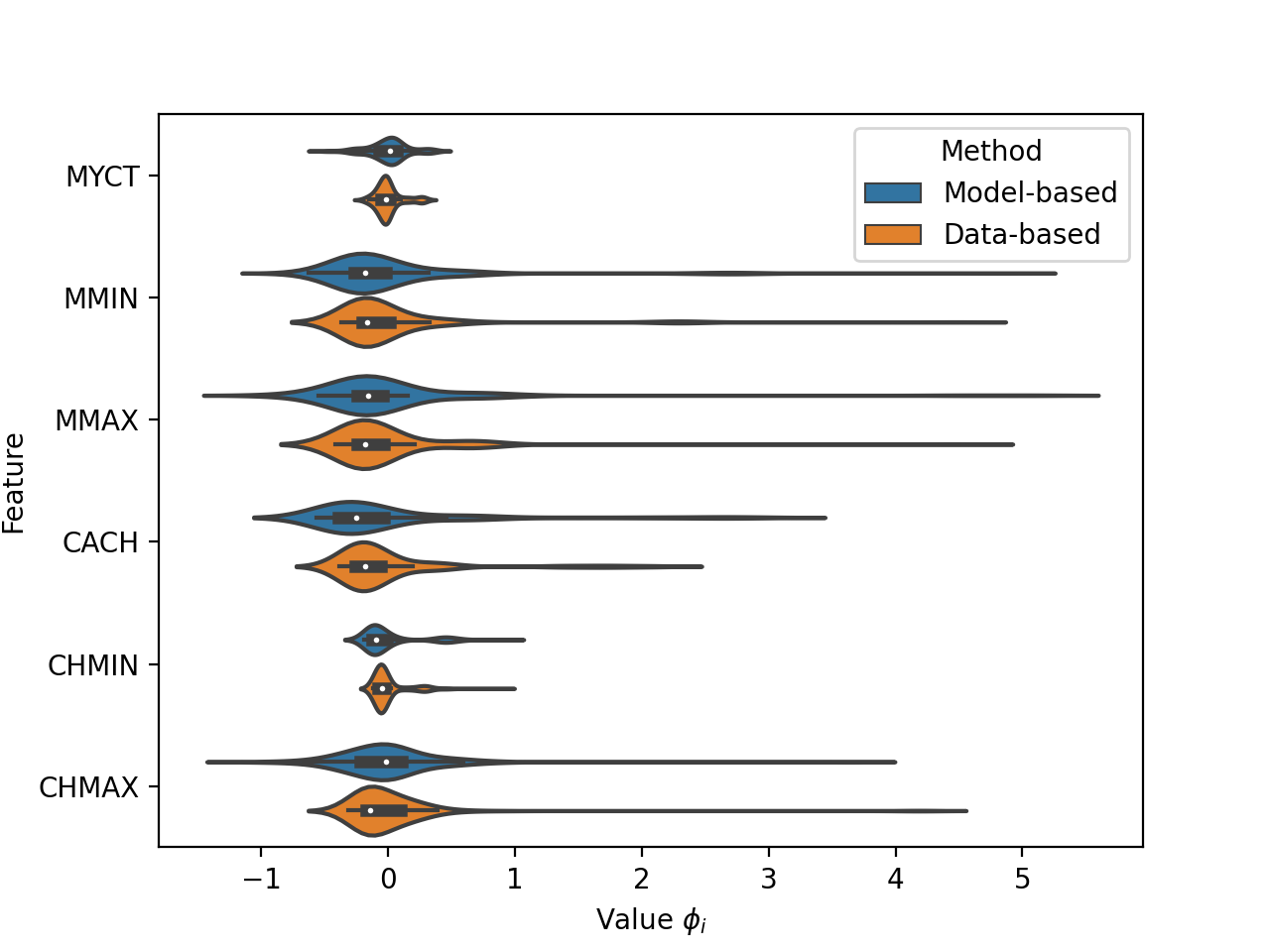}%
\caption{The violin plot of Shapley values obtained by using the data-based
SHAP (the first row of pictures) and the model-based SHAP (the second row of
pictures) for the dataset MachineCPU}%
\label{f:SHAP2}%
\end{center}
\end{figure}

Fig. \ref{f:SHAP2} depicts the violin plot of Shapley values for the same
dataset MachineCPU. It can be viewed as a statistics of Shapley values for the
whole dataset, including training and testing examples. One can see from the
plot that the largest Shapley value corresponds to the feature MMAX. Moreover,
we can conclude from Fig. \ref{f:SHAP2} that the model-based SHAP and the
data-based SHAP produce very similar Shapley values.%

Table \ref{t:SHAP-1} shows computational times in seconds for computing
Shapley values of $n$ examples by using the original SHAP, the data-based
SHAP, and the model-based SHAP. We use $10000$ iterations of GBM-HRBM under
condition that corners are used as base models. It can be seen from Table
\ref{t:SHAP-1} that computational time of the original SHAP linearly increases
with $n$ whereas the proposed modifications of SHAP are changed very slowly.%

\begin{table}[tbp] \centering
\caption{Computation time (seconds) of the original SHAP and two its GBM-HRBM modifications}%
\begin{tabular}
[c]{cccc}\hline
& \multicolumn{3}{c}{SHAP}\\ \hline
$n$ & Original & Data-based & Model-based\\ \hline
$1$ & $2.780$ & $0.874$ & $0.071$\\
$10$ & $27.130$ & $0.990$ & $0.073$\\
$20$ & $54.087$ & $1.062$ & $0.075$\\ \hline
\end{tabular}
\label{t:SHAP-1}%
\end{table}%

\begin{figure}
[ptb]
\begin{center}
\includegraphics[
height=2.962in,
width=5.7346in
]%
{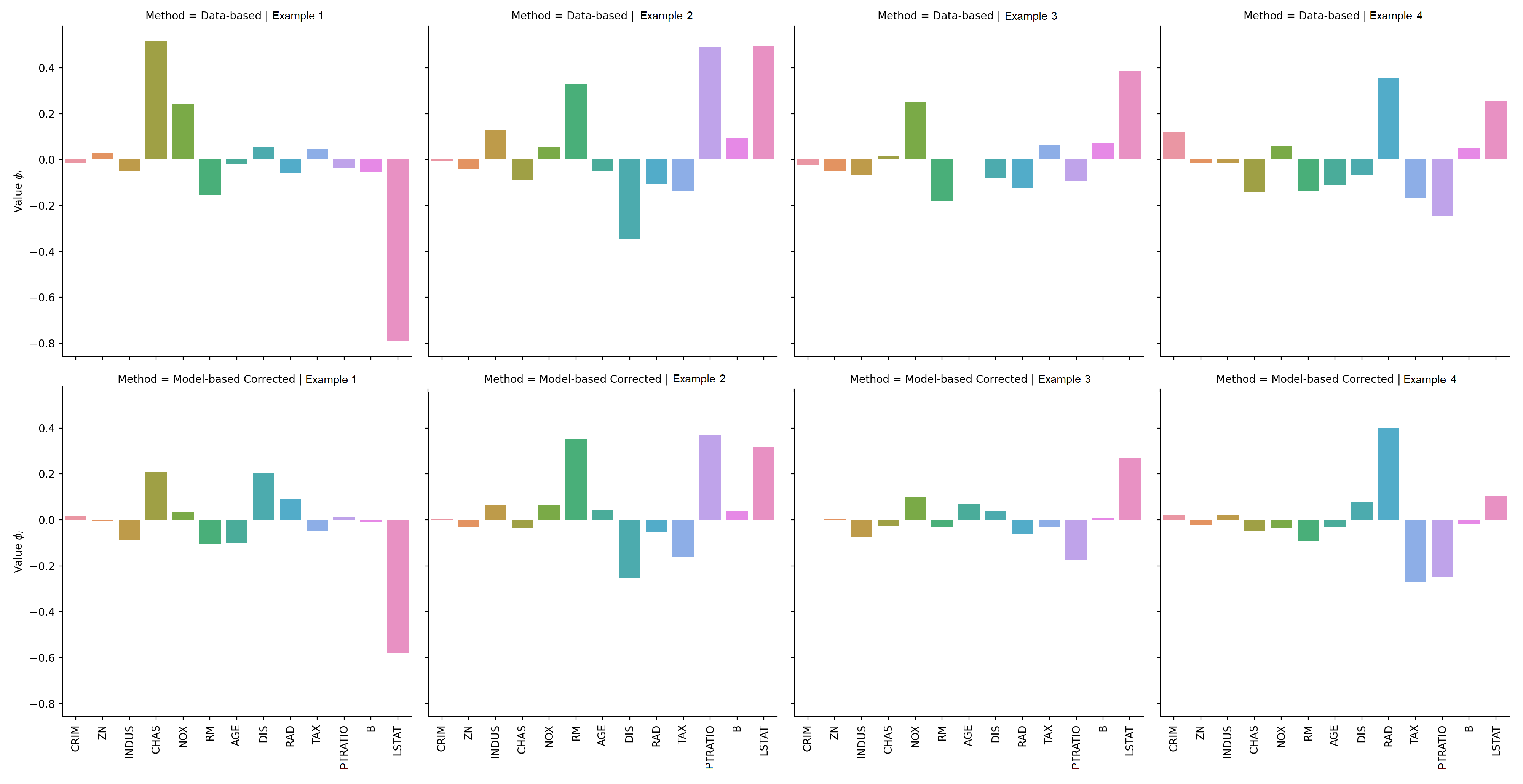}%
\caption{Comparison of Shapley values obtained by using the data-based SHAP
(the first row of pictures) and the model-based SHAP (the second row of
pictures) for the dataset Boston}%
\label{f:SHAP3}%
\end{center}
\end{figure}

\begin{figure}
[ptb]
\begin{center}
\includegraphics[
height=3.6867in,
width=2.5892in
]%
{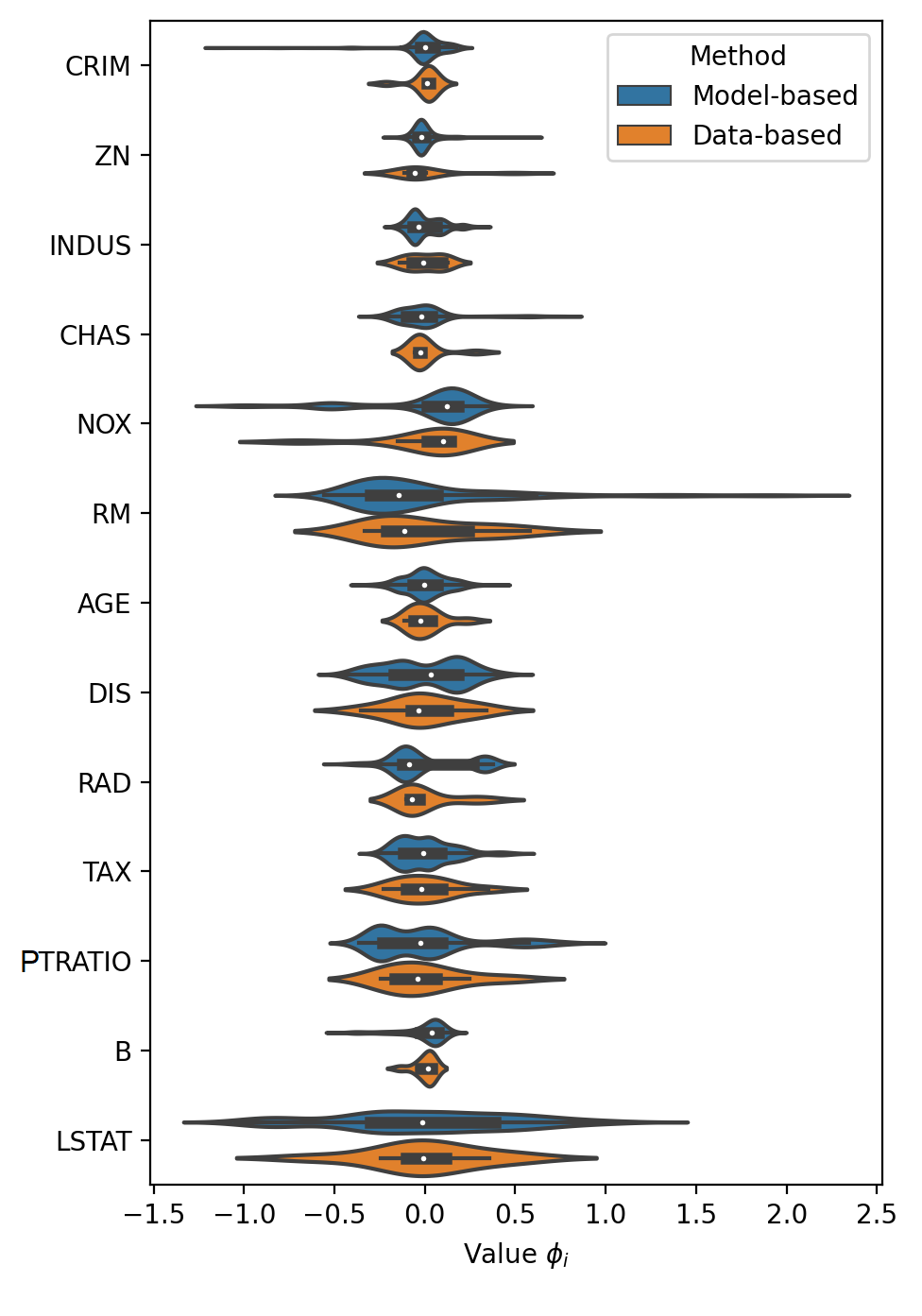}%
\caption{The violin plot of Shapley values obtained by using the data-based
SHAP (the first row of pictures) and the model-based SHAP (the second row of
pictures) for the dataset Boston}%
\label{f:SHAP4}%
\end{center}
\end{figure}

In order to see that the data-based and model-based SHAP models provide
similar Shapley values, we consider the dataset Boston. Fig. \ref{f:SHAP3}
shows Shapley values $\phi_{i}$ of all features (CRIM, ZN, INDUS, CHAS, NOX,
RM, AGE, DIS, RAD, TAX, PTRATIO, B, LSTAT), obtained by the same two methods.
We again randomly select four examples from the dataset for analyzing. Fig.
\ref{f:SHAP4} depicts the violin plot of Shapley values for the dataset
Boston. It can be seen from the plot that the largest Shapley value
corresponds to features RM and LSTAT. It is interesting to note that the same
features have been selected as the most important ones in
\cite{Konstantinov-Utkin-21}.%

%

\section{Conclusion}

A new ensemble model based on GBM with axis-parallel HRBMs as base models has
been proposed. Two types of HRBMs have been studied: closed rectangles and
corners. Numerical experiments have shown that corners mainly provide better
results in comparison with closed rectangles. The proposed ensemble-based
model is extremely simple due to simplicity of HRBMs. In spite of the model
simplicity, various numerical experiments with real data have demonstrated the
model efficiency and its overfitting prevention. If to compare HRBMs with
decision trees, which are the most popular base models in boosting methods,
then HRBM are much simpler than trees even with the minimal depth. At the same
time, GBM-HRBM outperforms GBM with decision trees.

Another advantage of GBM-HRBM is that optimal parameters of regularization can
be controlled at each iteration of GBM. This peculiarity of GBM-HRBM also
improves the proposed ensemble-based model and reduces the computational time
for training and testing the GBM-HRBM model.

A surprising peculiarity of the GBM-HRBM is that it is interpretable in a
simple way by applying the SHAP interpretation method. We do not need to get a
huge number of predictions for various subsets of features as it is performed
in the original SHAP. Moreover, we do not need to use one of the available
methods \cite{Covert-etal-21} for removing the features. Propositions
\ref{prop:rect_shap_1} and \ref{prop:rect_shap_2} significantly simplify SHAP
and provide accurate Shapley values.

It should be pointed out that HRBMs open a door for developing various
modifications which could improve and simplify the regression and
classification tasks. Therefore, several aspects of HRBMs will be covered in
subsequent works. First, algorithms for generating optimal rectangles, for
predicting and testing can be improved in order to reduce the computation
time. Second, expectations (\ref{SHAP-exp-est}) in the data-based modification
of SHAP can be calculated through estimate of the probability density function
based on the same rectangles by applying the well-known algorithms for the
density estimation. This is an interesting direction for further research.
Third, GBM-HRBM can be implemented as a simple and accurate interpretable
approximating meta-model for complex black-box models. Fourth, it is
interesting to study different algorithms to analyze training examples inside
rectangles, for instance, to apply the attention mechanism to correct the mean
target values of examples. It is expected that these algorithms will
significantly improve the whole GBM with HRBMs.

\bibliographystyle{unsrt}
\bibliography{Boosting,Classif_bib,Deep_Forest,Expl_Attention,Explain,Explain_med,MYBIB,MYUSE}

\appendix

\section{Appendix: Proof of Propositions}

\numberwithin{equation}{section} \setcounter{equation}{0}

\textbf{Proof of Proposition \ref{prop:regular1}}: If $\lambda_{1}=0$, then
the extended loss function $\tilde{\mathcal{L}}$ in (\ref{APR-40}) has a
minimum at a point with zero derivatives. This implies that there hold%
\begin{equation}
\dfrac{\partial \tilde{\mathcal{L}}}{\partial v_{in}}=\dfrac{1}{N}\left(
G_{in}+v_{in}H_{in}\right)  +\lambda_{2}v_{in}=0,
\end{equation}%
\begin{equation}
\dfrac{\partial \tilde{\mathcal{L}}}{\partial v_{out}}=\dfrac{1}{N}\left(
G_{out}+v_{out}H_{out}\right)  +\lambda_{2}v_{out}=0.
\end{equation}

Hence, we directly get (\ref{APR-30}). In the same way, the case when
$\lambda_{1}>0$ and $\lambda_{2}>0$ can be considered. In this case, we have
to take into account the sign of $v_{in}(\lambda_{1},\lambda_{2})$. Suppose
that $v_{in}(\lambda_{1},\lambda_{2})>0$. Then we write
\begin{equation}
\dfrac{\partial \tilde{\mathcal{L}}}{\partial v_{in}}=\dfrac{1}{N}\left(
G_{in}+v_{in}H_{in}\right)  +\lambda_{2}v_{in}+\lambda_{1}=0.
\end{equation}

Hence, there holds
\begin{equation}
v_{in}(\lambda_{1},\lambda_{2})=-\frac{G_{in}+N\cdot \lambda_{1}}{N\cdot
\lambda_{2}+H_{in}}. \label{APR-31}%
\end{equation}

It follows from (\ref{APR-31}) that condition $v_{in}(\lambda_{1},\lambda
_{2})>0$ is fulfilled when $G_{in}+N\cdot \lambda_{1}<0$ because the second
derivative $h_{i}$ is positive. In the same way, we can get other cases for
$v_{in}(\lambda_{1},\lambda_{2})$. Finally, we write (\ref{APR-32}).
Similarly, we can find values of $v_{out}(\lambda_{1},\lambda_{2})$.
$\blacksquare$

\textbf{Proof of Proposition \ref{prop:regular2}}: Note that there holds%
\begin{equation}
\frac{\partial \Omega_{h}}{\partial v_{in}}=-\frac{\partial \Omega_{h}}{\partial
v_{out}}.
\end{equation}

This implies that values of $v_{in}$ and $v_{out}$ are connected without a
dependence of the regularization parameters ($\eta_{1},\eta_{2}$) because the
following is valid:
\begin{equation}
0=\frac{\partial(\hat{L}+\Omega_{h})}{\partial v_{in}}+\frac{\partial(\hat
{L}+\Omega_{h})}{\partial v_{out}}=\frac{\partial \hat{L}}{\partial v_{in}%
}+\frac{\partial \hat{L}}{\partial v_{out}}.
\end{equation}

Hence, there holds
\begin{equation}
v_{out}=-\frac{G+v_{in}\cdot H_{in}}{H_{out}}. \label{APR-54}%
\end{equation}

If $\eta_{1}=0$, then we get (\ref{APR-52}). The value of $v_{out}(\eta_{2})$
can be obtained by substituting (\ref{APR-52}) into (\ref{APR-54}) and by
taking into account that there hold $G=G_{in}+G_{out}$ and $H=H_{in}+H_{out}$.

If $\eta_{1}>0$ and $\eta_{2}=0$, then we obtain
\begin{equation}
v_{in}(\eta_{1})=\left \{
\begin{array}
[c]{cc}%
-\frac{G_{in}-N\cdot \eta_{1}}{H_{in}}, & \frac{H_{in}G_{out}-H_{out}G_{in}}%
{H}<-N\cdot \eta_{1},\\
-\frac{G_{in}}{H_{in}}, & \left \vert \frac{H_{in}G_{out}-H_{out}G_{in}}%
{H}\right \vert \leq N\cdot \eta_{1},\\
-\frac{G_{in}+N\cdot \eta_{1}}{H_{in}}, & \frac{H_{in}G_{out}-H_{out}G_{in}}%
{H}>N\cdot \eta_{1}.
\end{array}
\right.
\end{equation}

It should be noted that $v_{in}(\eta_{1})=-G_{in}/H_{in}$ as well as
$v_{out}=-G_{out}/H_{out}$ do not depend on $\eta_{1}$ when $\left \vert
\frac{H_{in}G_{out}-H_{out}G_{in}}{H}\right \vert \leq N\cdot \eta_{1}$. This
implies that the regularization term is reduced to a constant. It is simply to
show that other cases of the regularization can be implemented by using only
the case $\eta_{2}>0$, $\eta_{1}=0$. $\blacksquare$

\textbf{Proof of Proposition \ref{prop:regular3}}: If $\mathbb{I}_{in}%
^{(i)}=1$, i.e., there holds $\mathbf{x}_{i}\in \mathbf{r}$, then
$\mathbb{I}_{out}^{(i)}=0$, and the condition (\ref{eq:abs_le_beta}) can be
rewritten by using (\ref{APR-20}) as
\begin{equation}
\left \vert v_{in}\right \vert \leq \beta. \label{APR-46}%
\end{equation}

If we substitute (\ref{APR-30}) into (\ref{APR-46}), then the constraint for
$v_{in}$ can be fulfilled by choosing the corresponding value of $\lambda_{2}$
in (\ref{APR-40}) denoted as $\lambda_{2}^{in}$:
\begin{equation}
\left \vert -\frac{G_{in}}{N\cdot \lambda_{2}^{in}+H_{in}}\right \vert \leq \beta.
\label{APR-47}%
\end{equation}

The constraint for $v_{out}$ can be similarly written as follows:%
\begin{equation}
\left \vert -\frac{G_{out}}{N\cdot \lambda_{2}^{out}+H_{out}}\right \vert
\leq \beta.
\end{equation}

Since the second derivative $h_{i}$ is positive for convex loss functions,
then it follows from (\ref{APR-47}) that there holds:
\begin{equation}
\underline{\lambda}_{2}^{in}=\frac{1}{N}\left[  \frac{1}{\beta}\left \vert
G_{in}\right \vert -H_{in}\right]  .
\end{equation}

Hence, the constraint (\ref{APR-46}) is fulfilled for all values $\lambda
_{2}^{in}\geq \underline{\lambda}_{2}^{in}$. The same can be written for
$\underline{\lambda}_{2}^{out}$. Hence, the maximum value of $\lambda_{2}$
defined as (\ref{APR-56})%
\begin{equation}
\lambda_{2}=\max \left(  {\underline{\lambda}_{2}^{in},\underline{\lambda}%
_{2}^{out},0}\right)
\end{equation}
satisfies both constraints and therefore (\ref{eq:abs_le_beta}).

Similarly, we can get parameters for the $L_{1}$ regularization by writing the
condition
\begin{equation}
\max \left(  \left \vert \frac{G_{in}+N\cdot \lambda_{1}}{H_{in}}\right \vert
,\left \vert \frac{G_{in}-N\cdot \lambda_{1}}{H_{in}}\right \vert \right)
\leq \beta.
\end{equation}

Hence, we obtain (\ref{APR-59})-(\ref{APR-59-2}). $\blacksquare$

\textbf{Proof of Proposition \ref{prop:regular4}}: First, we consider a case
when $\mathbb{I}_{in}^{(i)}=1$ and $f_{k}(\mathbf{x})=v_{in}(\eta_{2})$. Then
substituting (\ref{APR-52}) into (\ref{eq:abs_le_beta}) or (\ref{APR-46}), we
get the condition:
\begin{equation}
|v_{in}(\eta_{2})|=\left \vert \frac{G_{in}+N\cdot \eta_{2}\cdot \left(  \frac
{G}{H_{out}}\right)  }{H_{in}+N\cdot \eta_{2}\cdot \left(  \frac{H}{H_{out}%
}\right)  }\right \vert \leq \beta.
\end{equation}

It can be rewritten as the following two inequalities:%
\begin{equation}
-\beta \cdot \left(  H_{in}+\eta_{2}\cdot N\cdot \left(  \frac{H}{H_{out}%
}\right)  \right)  \leq G_{in}+\eta_{2}\cdot N\cdot \left(  \frac{G}{H_{out}%
}\right)  ,
\end{equation}%
\begin{equation}
\beta \cdot \left(  H_{in}+\eta_{2}\cdot N\cdot \left(  \frac{H}{H_{out}}\right)
\right)  \geq G_{in}+\eta_{2}\cdot N\cdot \left(  \frac{G}{H_{out}}\right)  .
\end{equation}

Hence, we get constraints for $\eta_{2}$:%
\begin{equation}
\eta_{2}\cdot \left(  G+\beta \cdot H\right)  \geq \frac{H_{out}}{N}(G_{in}%
+\beta \cdot H_{in}), \label{APR-70}%
\end{equation}%
\begin{equation}
\eta_{2}\cdot \left(  G-\beta \cdot H\right)  \leq-\frac{H_{out}}{N}%
(G_{in}+\beta \cdot H_{in}). \label{APR-71}%
\end{equation}

In order to find the minimal value $\eta_{2}$ which satisfies the above
constraints, three cases should be considered: $G>\beta H$, $-\beta H<G<\beta
H$, and $G<-\beta H$. Denote
\begin{equation}
C_{1}=\frac{H_{out}}{N}\frac{G_{in}+\beta H_{in}}{G+\beta H},\ C_{2}%
=-\frac{H_{out}}{N}\frac{G_{in}+\beta H_{in}}{G-\beta H}.
\end{equation}

\begin{enumerate}
\item If $G>\beta H$, then it follows from (\ref{APR-70}) and (\ref{APR-71})
that $\eta_{2}\geq C_{1}$ and $\eta_{2}\leq C_{2}$. Hence, the smallest
feasible solution for $\eta_{2}$ is $\underline{\eta}_{2}=C_{1}$ because the
second inequality provides non-negative values of $\eta_{2}$.

\item If $|G|<\beta H$, then $\eta_{2}\geq C_{1}$ and $\eta_{2}\geq C_{2}$.
Hence, the smallest feasible solution for $\eta_{2}$ is $\underline{\eta}%
_{2}=\max \left(  C_{2}{,}C_{1}\right)  $.

\item If $G<-\beta H$, then conditions for $\eta_{2}$ are $\eta_{2}\leq C_{1}$
and $\eta_{2}\geq C_{2}$. Hence, it is obvious that the smallest feasible
solution for $\eta_{2}$ is $\underline{\eta}_{2}=C_{2}$.
\end{enumerate}

Let us consider the second case when $\mathbb{I}_{out}^{(i)}=1$ and
$f_{k}(\mathbf{x})=v_{out}(\eta_{2})$. Since the condition
(\ref{eq:abs_le_beta}) has to be fulfilled for $v_{in}$ as well as for
$v_{out}$, then the minimal value of $\eta_{2}$ depends on the same three
cases. The corresponding bounds for $\eta_{2}$ can be obtained in the same
way. $\blacksquare$

\textbf{Proof of Proposition \ref{prop:rect_shap_1}}: Represent expression
(\ref{SHAP_2}) in another form:
\begin{equation}
\Psi(S)=\prod_{j\in S}\mathbb{I}[x^{(j)}\in r^{(j)}]\cdot(v_{in}%
-v_{out})+v_{out}. \label{phiS}%
\end{equation}

According to the fourth property of Shapley values (dummy), if the $j$-th
feature falls in the $j$-th coordinate of the rectangle, i.e., $\mathbb{I}%
[x^{(j)}\in r^{(j)}]=1$, then its contribution is zero. Let the observation be
inside the rectangle by means of the $i$-th feature, i.e., there holds
\begin{equation}
\mathbb{I}[x^{(i)}\in r^{(i)}]=1.
\end{equation}

The above can be proved as follows. First, we can write
\begin{align}
\Psi(S\cup \{i\})  &  =\mathbb{I}[x^{(i)}\in r^{(i)}]\cdot \prod_{j\in
S}\mathbb{I}[x^{(j)}\in r^{(j)}]\cdot(v_{in}-v_{out})+v_{out}\nonumber \\
&  =\Psi(S).
\end{align}

Hence, there holds
\begin{equation}
\phi_{i}=0.
\end{equation}

For all other features, the second property (symmetry) holds in pairs. Indeed,
if we write
\begin{equation}
\mathbb{I}[x^{(j)}\in r^{(j)}]=0,\  \mathbb{I}[x^{(k)}\in r^{(k)}]=0,
\label{eq:IjIk0}%
\end{equation}
then the following is correct:
\begin{equation}
\Psi(S\cup \{j\})=v_{out}=\Psi(S\cup \{k\}).
\end{equation}

Hence, there holds%
\begin{equation}
\phi_{j}=\phi_{k}.
\end{equation}

It follows from the first property (efficiency) that
\begin{equation}
\Psi(\{1,\dots,d\})=\Psi(\emptyset)+\sum_{i=1}^{d}\phi_{i}=\Psi(\emptyset
)+\sum_{k=1}^{u}\tilde{\phi}=\Psi(\emptyset)+u\cdot \tilde{\phi},
\end{equation}
where $u$ is the number of features for which the observation is outside
$\mathbf{r}$; $\tilde{\phi}$ is a value of contribution for these features.

Then we get%
\begin{equation}
\phi_{i}=%
\begin{cases}
0, & x^{(i)}\in r^{(j)},\\
\dfrac{\Psi(\{1,\dots,d\})-\Psi(\emptyset)}{\sum_{j=1}^{d}\mathbb{I}%
[x^{(i)}\notin r^{(j)}]}, & x^{(i)}\notin r^{(j)},
\end{cases}
\end{equation}
or, taking into account (\ref{phiS}) and (\ref{phi0}), we obtain
(\ref{SHAP_3}), as was to be proved. $\blacksquare$

\textbf{Proof of Proposition \ref{prop:rect_shap_2}}: Since the distribution
of data is unknown, then the expectation of model predictions can be
calculated by using some estimate of the density. Another way is to
approximate the average over the sample. Due to simplicity, the second way is
more preferable. Moreover, it does not introduce an additional error caused by
a density estimation method. Hence, we write%
\begin{equation}
\tilde{\Psi}(\emptyset)\approx v_{out}+(v_{in}-v_{out})\cdot \frac{1}{N}%
\sum_{t=1}^{N}\mathbb{I}[\mathbf{x}_{t}\in \mathbf{r}].
\end{equation}

To calculate the subset function value, we have to estimate $\tilde{\Psi}(S)$
as
\begin{equation}
\tilde{\Psi}(S)\approx v_{out}+(v_{in}-v_{out})\cdot \mathbb{E}\left[
\prod_{i=1}^{d}\mathbb{I}[X_{i}\in r^{(i)}]\mid X_{S}=\mathbf{x}^{(S)}\right]
,
\end{equation}
where the expectation can be represented as:
\begin{equation}
\prod_{i\in S}\mathbb{I}[x_{i}\in r_{i}]\cdot \mathbb{E}\left[  \prod_{i\notin
S}\mathbb{I}[X_{i}\in r_{i}]\mid X_{S}=\mathbf{x}^{(S)}\right]  .
\end{equation}

To estimate given expectations, the assumption is introduced
\cite{Lundberg-Lee-2017,Strumbel-Kononenko-2010} that the feature random
variables $S$ and $\overline{S}$ are independent. In this case, we can write
\begin{align}
&  \mathbb{E}\left[  \mathbb{I}[X_{\overline{S}}\in \mathbf{r}^{(\overline{S}%
)}]\mid X_{S}=\mathbf{x}^{(S)}\right] \nonumber \\
&  =\mathbb{E}_{X_{\overline{S}}}\left[  \mathbb{I}[X_{\overline{S}}%
\in \mathbf{r}^{(\overline{S})}]\right]  \approx \frac{1}{N}\sum_{t=1}%
^{N}\mathbb{I}[\mathbf{x}_{t}^{(\overline{S})}\in \mathbf{r}^{(\overline{S})}].
\end{align}

Thus, accepting a strong assumption about the independence of feature subsets,
we finally get:
\begin{equation}
\tilde{\Psi}(S)=v_{out}+(v_{in}-v_{out})\mathbb{I}[\mathbf{x}^{(S)}%
\in \mathbf{r}^{(S)}]\cdot \frac{1}{N}\sum_{t=1}^{N}\mathbb{I}[\mathbf{x}%
_{t}^{(\overline{S})}\in \mathbf{r}^{(\overline{S})}].
\end{equation}

Similarly to the first method (Proposition \ref{prop:rect_shap_1}), we use the
symmetry property. If (\ref{eq:IjIk0}) is valid, then there holds
\begin{equation}
\tilde{\Psi}(S\cup \{j\})=v_{out}=\tilde{\Psi}(S\cup \{k\}),
\end{equation}
and, therefore, the equality $\phi_{j}=\phi_{k}=\widetilde{\phi}$ is valid.
Otherwise, if the observation is inside the rectangle by feature $i$, then:
\begin{align}
\tilde{\Psi}(S\cup \{i\})  &  =v_{out}+(v_{in}-v_{out})\cdot1\cdot
\mathbb{I}[\mathbf{x}^{(S)}\in \mathbf{r}^{(S)}]\nonumber \\
\times \frac{1}{N}\sum_{t=1}^{N}\mathbb{I}[\mathbf{x}_{t}^{(\overline
{S}\setminus \{i\})}  &  \in \mathbf{r}^{(\overline{S}\setminus \{i\})}].
\end{align}

In sum, we get (\ref{SHAP_6})-(\ref{SHAP_8}). $\blacksquare$
\end{document}